\documentclass[10pt,twocolumn,letterpaper]{article}
\pdfoutput=1
\usepackage{wacv}
\usepackage{times}
\usepackage{epsfig}
\usepackage{graphicx}
\usepackage{amsmath}
\usepackage{amssymb}

\usepackage{multirow}
\usepackage{stmaryrd}
\usepackage{arydshln}
\usepackage{pifont}
\usepackage{threeparttable, tablefootnote}

\wacvfinalcopy 

\ifwacvfinal
\def\assignedStartPage{1} 
\fi

\ifwacvfinal
\usepackage[breaklinks=true,bookmarks=false]{hyperref}
\else
\usepackage[pagebackref=true,breaklinks=true,colorlinks,bookmarks=false]{hyperref}
\fi

\usepackage{hyperref}
\hypersetup{
    colorlinks=true,
    linkcolor=blue,
    filecolor=magenta,      
    urlcolor=magenta,
}

\ifwacvfinal
\else
\fi

\begin{document}

\title{V-SlowFast Network for Efficient Visual Sound Separation}

\author{Lingyu Zhu\\
Computer Vision Group\\
Tampere University, Finland\\
{\tt\small lingyu.zhu@tuni.fi}
\and
Esa Rahtu\\
Computer Vision Group\\
Tampere University, Finland\\
{\tt\small esa.rahtu@tuni.fi}
}

\maketitle

\begin{abstract}
    The objective of this paper is to perform visual sound separation: i) we study visual sound separation on spectrograms of different temporal resolutions; ii) we propose a new light yet efficient three-stream framework $\textbf{V-SlowFast}$ that operates on $\textbf{V}$isual frame, $\textbf{Slow}$ spectrogram, and $\textbf{Fast}$ spectrogram. The $\textbf{Slow}$ spectrogram captures the coarse temporal resolution while the $\textbf{Fast}$ spectrogram contains the fine-grained temporal resolution; iii) we introduce two contrastive objectives to encourage the network to learn discriminative visual features for separating sounds; iv) we propose an audio-visual global attention module for audio and visual feature fusion; v) the introduced $\textbf{V-SlowFast}$ model outperforms previous state-of-the-art in single-frame based visual sound separation on small- and large-scale datasets: MUSIC-21, AVE, and VGG-Sound. We also propose a small $\textbf{V-SlowFast}$ architecture variant, which achieves 74.2\% reduction in the number of model parameters and 81.4\% reduction in GMACs compared to the previous multi-stage models. Project page: \href{https://ly-zhu.github.io/V-SlowFast}{https://ly-zhu.github.io/V-SlowFast}. 

\end{abstract}



\section{Introduction}
\label{sec:intro}

Sound source separation aims at extracting the target source from a given audio mixture. The audio-based source separation task~\cite{ghahramani1996factorial,roweis2001one,virtanen2007monaural,cichocki2009nonnegative} has been extensively studied in the audio processing community. However, the task remains challenging due to the underdetermined nature of source separation problem. The cocktail party problem~\cite{golumbic2013visual,ephrat2018looking} is a well known example, where one attempts to follow one of the discussions while multiple people are talking simultaneously.

Recent works~\cite{Zhao_2018_ECCV,ephrat2018looking,Zhao_2019_ICCV,gao2019co,zhu2020visually,zhu2021leveraging,gan2020music,zhu2021visually} have started to exploit visual information (e.g. talking face, playing instruments) to solve the sound separation task. For instance, visual cues like object categories or movements can be used to facilitate the source separation problem. While visual motions may be important under certain circumstances (e.g. separating similar type of sources), the single visual frame based approaches have demonstrated surprisingly well performance  in~\cite{Zhao_2018_ECCV,zhu2020visually,zhu2021leveraging}. In this paper, we focus on improving the single visual frame based sound separation.

\begin{figure}[t]
    \centering
    \begin{tabular}{c}
        \includegraphics[width=1.0\linewidth]{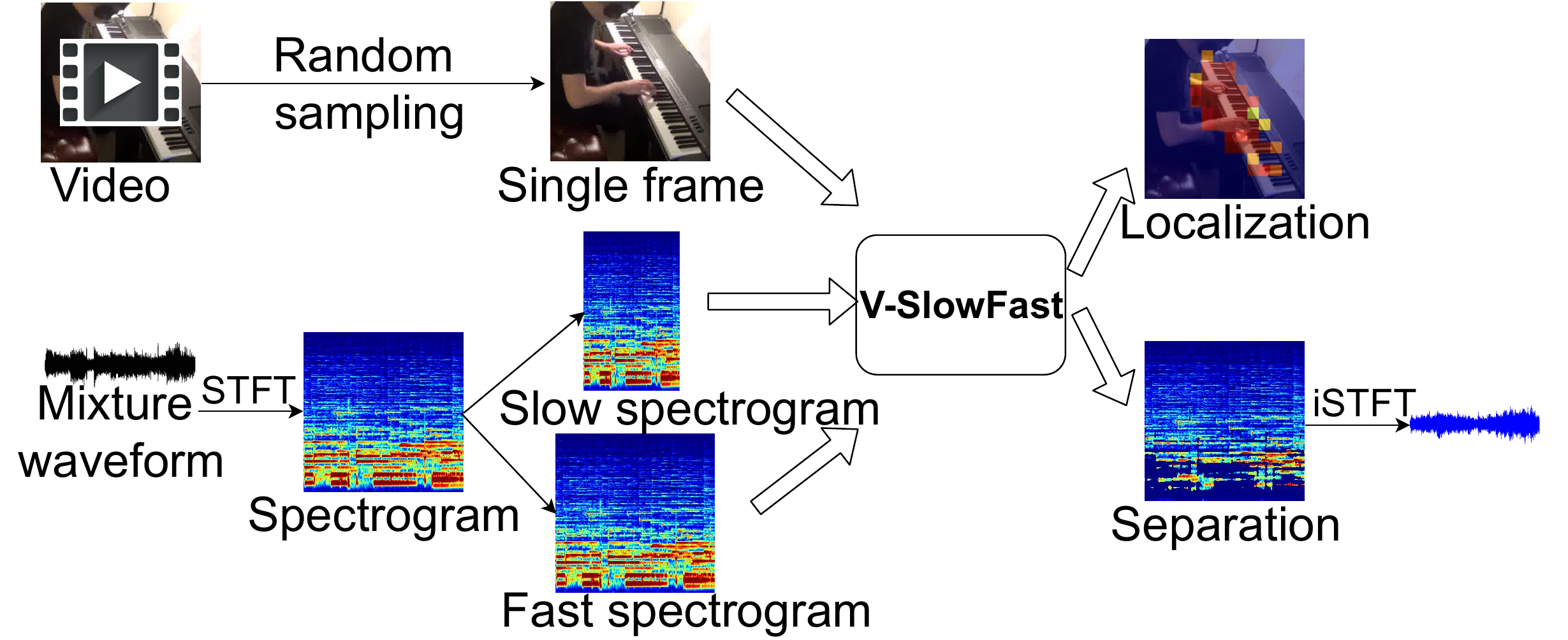}
    \end{tabular}
    \caption{The $\textbf{V-SlowFast}$ network takes a video and mixture waveform as input. It operates on a visual frame (extracted from video), slow spectrogram (low temporal resolution), and fast spectrogram (high temporal resolution). Eventually, the V-SlowFast model can efficiently separate and localize sound sources.}
    \label{fig:fig1}
\end{figure}

Natural sounds have wide range of rhythms. For example, slow attacks or fast tweaks occur fairly frequently when someone plays an instrument. In order to gain a new perspective on perceiving natural sounds in the sound separation task, we implement a system to treat the slow attacks and fast tweaks separately. The concept of slow-fast networks have shown impressive success in video~\cite{feichtenhofer2019slowfast,xiao2020audiovisual} and audio~\cite{kazakos2021slow} recognition tasks, which operate on two streams of video frames or audio spectrograms with different sampling rates. Differently, we propose a novel three-stream framework $\textbf{V-SlowFast}$ (Figure~\ref{fig:fig1}) for the visually guided sound separation task: $\textbf{V}$ision, $\textbf{Slow}$, and $\textbf{Fast}$ pathway operating on visual frame, slow spectrogram, and fast spectrogram, respectively. The $\textbf{Slow}$ spectrogram pathway has coarse temporal resolution (low sampling rate) while the $\textbf{Fast}$ spectrogram pathway operates at fine-grained temporal resolution (high sampling rate). Moreover, we apply the concept of contrastive learning to the vision network for gaining discriminative semantic representations, which provide categorical cues (e.g. instrument type) for separating sounds and localizing sounding sources. Furthermore, we introduce an audio-visual global attention module (AVGA) to fuse the audio and visual features for making the model concentrate on the target sound source by leveraging corresponding global visual attention. Next, we upsample the global attended spectrum features to predict a mask for separating each component audio from mixture.



Multi-stage architectures~\cite{Xu_2019_ICCV,zhu2020visually,zhu2021visually} have shown good performance on visual source separation. However, these models tend to be large with high computational costs. We examine multiple options based on combinations of different spectrogram pathways (on different temporal resolutions) and different network architecture variants. To discover this, the \textbf{V-SlowFast} operates on spectrograms in multiple temporal resolutions. This is in contrast to previous works (e.g.~\cite{Xu_2019_ICCV,zhu2020visually}), where the separation is done only based on the spectrogram with full temporal resolution at each stage. On the one hand, we show that the \textbf{V-SlowFast} network can greatly improve the sound separation performance (SDR: 10.89) over the recent single visual frame based multi-stage system~\cite{zhu2020visually} (SDR: 9.50) and recursive method~\cite{Xu_2019_ICCV} (SDR: 9.15). On the other hand, we also propose a small \textbf{V-SlowFast} architecture variant, which contains only 15M parameters and consumes 0.84 GMACs~\cite{mcquillanfeasibility} for achieving similar result as previous multi-stage and recursive models (e.g. 58M parameters and 4$\sim$5 GMACs).

\section{Related Work}
\label{sec:relate}

\paragraph{\bf Audio-Visual Learning}
Audio-visual learning combines signals from different modalities: audio and vision. Recent works~\cite{aytar2016soundnet,arandjelovic2017look,arandjelovic2018objects} associate the learnt audio and visual embeddings by leveraging their correspondence. Synchronization based cross-modal approaches~\cite{Owens_2018_ECCV,korbar2018cooperative,cheng2020look,zhu2021visually} are proposed for visual representation learning. Another interesting task is to localize objects that sound~\cite{Zhao_2018_ECCV,arandjelovic2018objects,tian2018audio,zhu2020visually,zhu2021visually,chen2021localizing}, where the goal is to pinpoint audio sources from the visual data. Other interesting works study audio-visual action recognition~\cite{kazakos2019epic,korbar2019scsampler,gao2020listen,wu2021exploring}, audio-visual navigation~\cite{gao2020visualechoes,chen2020soundspaces,chen2021semantic}, talking head synthesis~\cite{wang2021one}, spatial audio from video~\cite{morgado2018self,gao20192,yang2020telling,morgado2020learning}, and visual-to-auditory~\cite{hu2019listen,gan2020foley}.


\paragraph{\bf Visual Sound Separation}
Early work~\cite{barzelay2007harmony} performs audio-visual sound attribution by leveraging the tight associations between audio and visual onset signal. Recently, Zhao~\textit{et al.}~\cite{Zhao_2018_ECCV,Zhao_2019_ICCV} proposed pioneering works to utilize appearance and motion cues for separating sound sources. Gao~\textit{et al.}~\cite{gao2019co,gao2018learning} studied to use object detection to facilitate source separation. Xu~\textit{et al.}~\cite{Xu_2019_ICCV} proposed a recursive model for separating sounds. Zhu~\textit{et al.}~\cite{zhu2020visually} further improved the models by utilizing visual cues of all the opponent sources. Gan~\textit{et al.}~\cite{gan2020music} associated keypoint-based body and finger movements with audio signals to separate sound sources. Owens~\textit{et al.}~\cite{Owens_2018_ECCV} and Zhu~\textit{et al.}~\cite{zhu2021visually} proposed synchronization based approaches for source separation. These works demonstrated how semantic appearances and motions could be utilized for sound separation. However, these works solely use full resolution spectrogram, which often leads to unnecessarily complex models.



\paragraph{\bf Self-Supervised Contrastive Learning}
Contrastive learning leverages multiple perspectives of the data to learn discriminative features. It has been actively studied recently for images~\cite{oord2018representation,caron2018deep,zhu2021leveraging,he2020momentum,chen2020simple,tian2020contrastive}, videos~\cite{wang2019learning,xu2019self,jenni2020video}, text~\cite{sun2019learning,alayrac2020self}, optical-flow~\cite{han2020self,tian2020contrastive}, and audio-video~\cite{patrick2020multi,piergiovanni2020evolving,alwassel_2020_xdc,ma2021active,zhu2021visually,ma2021contrastive}. In relation to previous efforts, our work studies two visual feature based contrastive objectives to obtain discriminative visual representation for visual sound separation and localization.


\paragraph{\bf SlowFast Networks}
There is a classical branch of works focusing on the two-stream methods~\cite{simonyan2014two,feichtenhofer2016convolutional,Carreira_2017_CVPR}, which exploit two different stream modalities (e.g. RGB images and flow). Recently, Feichtenhofer~\textit{et al.}~\cite{feichtenhofer2019slowfast} introduced a SlowFast network, which contains two pathways separately working at low and high framerates for video recognition. Similarly, Kazakos~\textit{et al.}~\cite{kazakos2021slow} proposed a two-stream convolutional network for audio recognition, that operates on low and high time-frequency spectrogram inputs. Xiao~\textit{et al.}~\cite{xiao2020audiovisual} proposed slow and fast visual pathways that are integrated with a faster audio pathway to model vision and sound for video recognition. Inspired from previous research in multimodel and multi-resolution models, we propose a novel three-stream framework $\textbf{V-SlowFast}$ for the visual sound separation task, which operates on visual frame and  spectrograms of slow and fast sampling rates.






\section{Approach}
\label{sec:method}

In this section, we first give a brief overview of our system (Sec.~\ref{sec:overview}). Then we propose a novel $\textbf{V-SlowFast}$ network for visual sound separation, that associates $\textbf{V}$ision  (Sec.~\ref{sec:vision_net}), spectrogram of $\textbf{Slow}$ sampling rate (Sec.~\ref{sec:slow}), and spectrogram of $\textbf{Fast}$ sampling rate (Sec.~\ref{sec:fast_residual}). Finally, we present our learning objective in Section~\ref{sec:objective}.



\subsection{Overview}
\label{sec:overview}

\begin{figure*}[t]
    \centering
    \begin{tabular}{c}
        \includegraphics[width=0.95\linewidth]{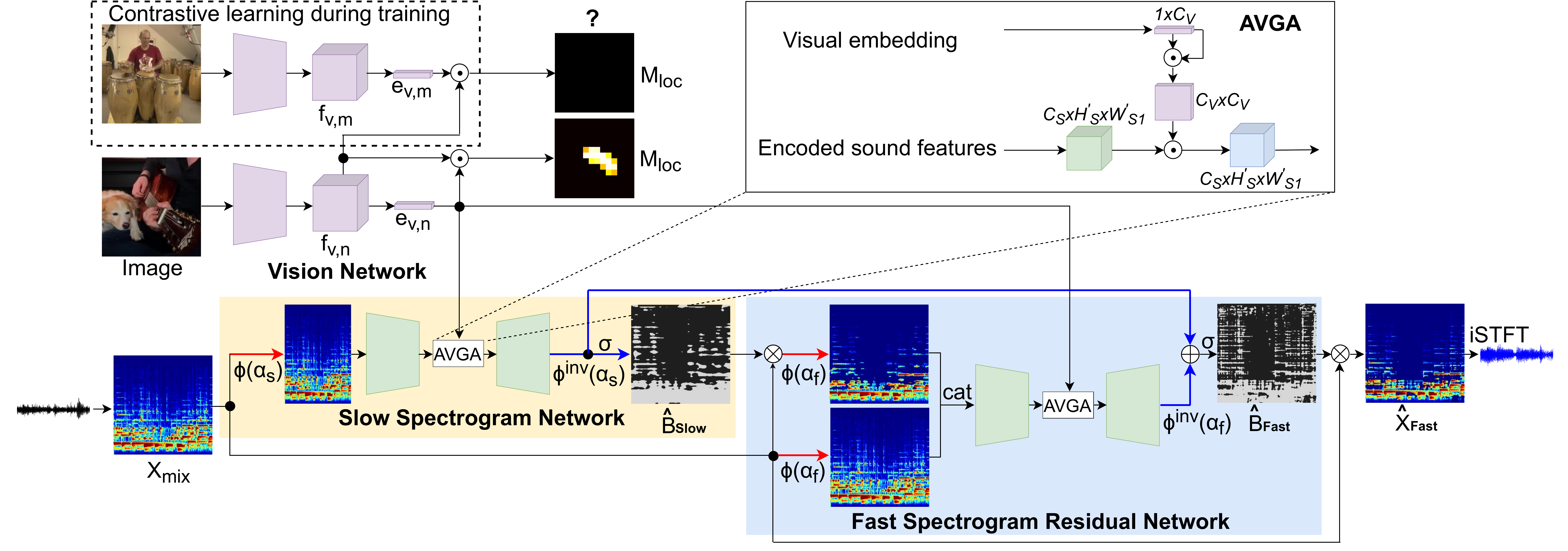}
    \end{tabular}
    \caption{The overview of the proposed V-SlowFast framework for visual sound separation. It contains four components: vision network, audio-visual global attention (AVGA) module, slow spectrogram network, and fast spectrogram residual network. The slow and fast network operate on low and high temporal resolution spectrograms respectively. Note that the contrastive learning (dashed block) is only applied during training.}
    \label{fig:overview}
\end{figure*}


The goal of the visual sound separation is to extract the component audio that corresponds to the sound source in the given visual frame. Figure~\ref{fig:overview} illustrates the overall architecture of the proposed $\textbf{V-SlowFast}$ network, which contains four components: vision network, audio-visual global attention module, slow spectrogram network, and fast spectrogram residual network. The vision network randomly extracts a single frame from the input video sequence and encodes it into a feature vector. To enhance the discrimination between semantic categories, we randomly sample an additional visual frame from a same (positive) or different (negative) category video to make contrastive pairs during the training procedure. We apply two visual contrastive objectives (embedding and localization) to the contrastive pairs along the vision network. The audio-visual global attention module fuses the visual embedding with sound features. The slow spectrogram network performs source separation at the coarse time scale (low sampling rate) using appearance features. The obtained result and the original mixture are further passed to the fast spectrogram residual network, which refines the source separation using spectrogram with higher temporal resolution (high sampling rate). The following sections provide further details of the system components and the learning objectives.

\subsection{Vision Network}
\label{sec:vision_net}

The Vision Network $V$ receives a randomly sampled frame $I$ from the input video and applies a dilated Res18-2D~\cite{he2016deep} or MV2~\cite{sandler2018mobilenetv2} to obtain a semantic representation $e_{v}$. More specifically, given an input RGB image $I \in \mathit{\mathbb{R}^{3 \times H_{V} \times W_{V}}}$, the Vision Network produces feature maps $f_{v} \in \mathit{\mathbb{R}^{C_{V} \times H_{V}^{'} \times W_{V}^{'}}}$. These are passed to a spatial average pooling layer to obtain visual embedding $e_{v} \in \mathit{\mathbb{R}^{1 \times C_{V}}}$,
\begin{equation}
    \centering
        \mathit{f_{v} = V(I), \quad e_{v}} = \mathit{spatial\_pool(f_{v})},
    \label{eq:vn}    
\end{equation}
where the $\mathit{C_{V}}$ denotes the dimension of visual features. $\mathit{H_{V}^{'} = H_{V} / 16}$ and $\mathit{W_{V}^{'} = W_{V} / 16}$.

\paragraph{\bf Visual Contrastive Learning Objectives}


We introduce two visual contrastive learning objectives: $\mathcal{L}_{e}(m, n, y)$ (embedding) and $\mathcal{L}_{M}(m, n, y)$ (localization) to the vision network. For the corresponding visual frame of each source $n$, we randomly sample an additional visual frame from a same or different category video $m$ to form contrastive learning pair. $y = 0$ (negative) if $m, n$ are of different type and $1$ (positive) otherwise. We formulate the visual embedding contrastive learning objective $\mathcal{L}_{e}(m, n, y)$ to enforce large visual embedding distances between negative pair and small distances for positive pair. The inner product between the visual features $f_{v,n}$ and visual embedding $e_{v,m}$ yields a sounding source location mask $M_{loc}(m,n)$. Therefore, we define a localization contrastive objective $\mathcal{L}_{M}(m, n, y)$, with a binary cross entropy (BCE) loss between the location mask $M_{loc}(m,n)$ and $y$, to enforce empty localization mask between negative pair and non-empty mask of sounding objects for positive pair. More specifically,
\begin{equation}
    \centering
    \begin{split}
        \mathit{dist(m, n)} = & \sum {\big(e_{v, m} - e_{v, n}\big)}^{2}, \\
        \mathit{\mathcal{L}_{e}(m, n, y)} = & \frac{1}{2} y \cdot \mathit{dist(m, n)} + \frac{1}{2} (1 - y) \cdot \\ \mathit{max}\{0, \mathit{margin} &- \sqrt{\mathit{dist}(m, n) + e^{-9}}\}^{2}, \\
        \mathit{M_{loc}}(m, n) = & \mathit{pool}\big(\sigma (e_{v,m} \odot f_{v,n})\big), \\
        \mathit{\mathcal{L}_{M}(m, n, y)} = &\mathit{BCE(M_{loc}(m, n), y)}, \\
        \mathit{\mathcal{L}_{contrast}} = & r_{1} \cdot \mathcal{L}_{e}(m, n, y) + r_{2} \cdot \mathcal{L}_{M}(m, n, y), 
    \end{split}
    \label{eq:contrastAA}    
\end{equation}
where the $\mathit{dist(m, n)}$ indicates the visual embedding distance between source $m$ and $n$. We adopt $\mathit{margin}=1.0$. A scalar product between the semantic embedding $e_{v,m}$ and the visual features $f_{v,n}$ results in a location mask $\mathit{M_{loc}}(m, n)$. $\sigma$ and $pool$ represent the $\mathit{sigmoid}$ and $\mathit{max\_pool}$ operation, respectively. $r_{1}$=$r_{2}$=$0.1$ control the contribution of each objective factor.

\subsection{Slow Spectrogram Network}
\label{sec:slow}

We adopt an encoder-decoder style architecture of U-Net~\cite{ronneberger2015u} or DeepLabV3Plus~\cite{chen2018encoder} for the slow spectrogram network. The U-Net consists of 7 down- and 7 up-convolutional layers with skip connection followed by a BatchNorm layer and a Leaky ReLU. MobileNetV2 (MV2)~\cite{sandler2018mobilenetv2} is adopted as the backbone of the DeepLabV3Plus. The input of the slow spectrogram network is a 2D frequency-time spectrogram of mixture sound and the output is a same-size binary spectrogram mask.

\paragraph{\bf Encoder}
The original audio mixture waveform is converted to a spectrogram presentation $X_{mix} \in \mathit{\mathbb{R}^{1 \times H_{S} \times W_{S}}}$ using Short-time Fourier Transform (STFT). We downsample the 2D frequency-time spectrogram $X_{mix}$, with a downsampling rate of $\alpha_{s}$ along the temporal dimension, as $\phi(X_{mix}, \alpha_{s})$. The encoder $\mathit{Slow^{E}}$ takes the mixture spectrogram $\phi(X_{mix}, \alpha_{s})$ as input and produces a feature representation $\mathit{f_{Slow^{E}, mix, \alpha_{s}}}$, as follows,
\begin{equation}
    \centering
    \mathit{f_{Slow^{E}, mix, \alpha_{s}} = Slow^{E}\big(\phi(X_{mix}, \alpha_{s})\big), \quad \alpha_{s} > 1},
    \label{eq:ss_encoder}    
\end{equation}
where the $\phi$ represents the temporal downsampling operation, $\alpha_{s}$ indicates the downsampling rate for the slow spectrogram network. $\phi(X_{mix}, \alpha_{s}) \in \mathit{\mathbb{R}^{1 \times H_{S} \times W_{S1}}}$ and $\mathit{f_{Slow^{E}, mix, \alpha_{s}} \in \mathbb{R}^{C_{S} \times H_{S}^{'} \times W_{S1}^{'}}}$. $\mathit{C_{S}}$, $\mathit{H_{S}}$ and $\mathit{W_{S}}$ denote the dimension of sound features, and frequency-time bases of the sound spectrogram, respectively. The $\mathit{W_{S1} = W_{S} / \alpha_{s}}$, $\mathit{H_{S}^{'} = H_{S} / 16}$ and $\mathit{W_{S1}^{'} = W_{S1} / 16}$.

\paragraph{\bf Audio-Visual Global Attention module} Previous works \cite{Owens_2018_ECCV,Zhao_2019_ICCV,gan2020music,zhu2021visually} have exploited the early fusion of the visual and audio features. However, these methods are mainly designed for fusing the visual motions into the sound features. In this paper, we propose an audio-visual global attention module (AVGA) in Figure~\ref{fig:overview} to fuse the semantic embedding $e_{v}$ into the middle part of the sound spectrogram network for making the model concentrate on the target source by leveraging corresponding global visual attention. In order to keep the channel and spatial dimension of the sound features unchanged, we adopt a self attention to the semantic embedding before fusing with the sound features. 
\begin{equation}
    \centering
    \mathit{f_{Slow^{E,V}, n, \alpha_{s}} = e_{v,n}^{T} \odot e_{v,n} \odot f_{Slow^{E}, mix, \alpha_{s}}},
    \label{eq:avga}    
\end{equation}
where the $\odot$ represents scalar product. The visual features $e_{v,n}$ of $n$-th source first have a self-attention with its own transposed embedding, then multiply with the encoded slow mixture sound features for providing global categorical attention. $\mathit{f_{Slow^{E,V}, n, \alpha_{s}}} \in \mathit{\mathbb{R}^{C_{S} \times H_{S}^{'} \times W_{S1}^{'}}}$ denotes the $n$-th global visual-attended sound features. Note that $C_{S}$ equals to the $C_{V}$ in Section~\ref{sec:vision_net}.

\begin{figure}[t]
    \centering
    \begin{tabular}{c}
        \includegraphics[width=0.95\linewidth]{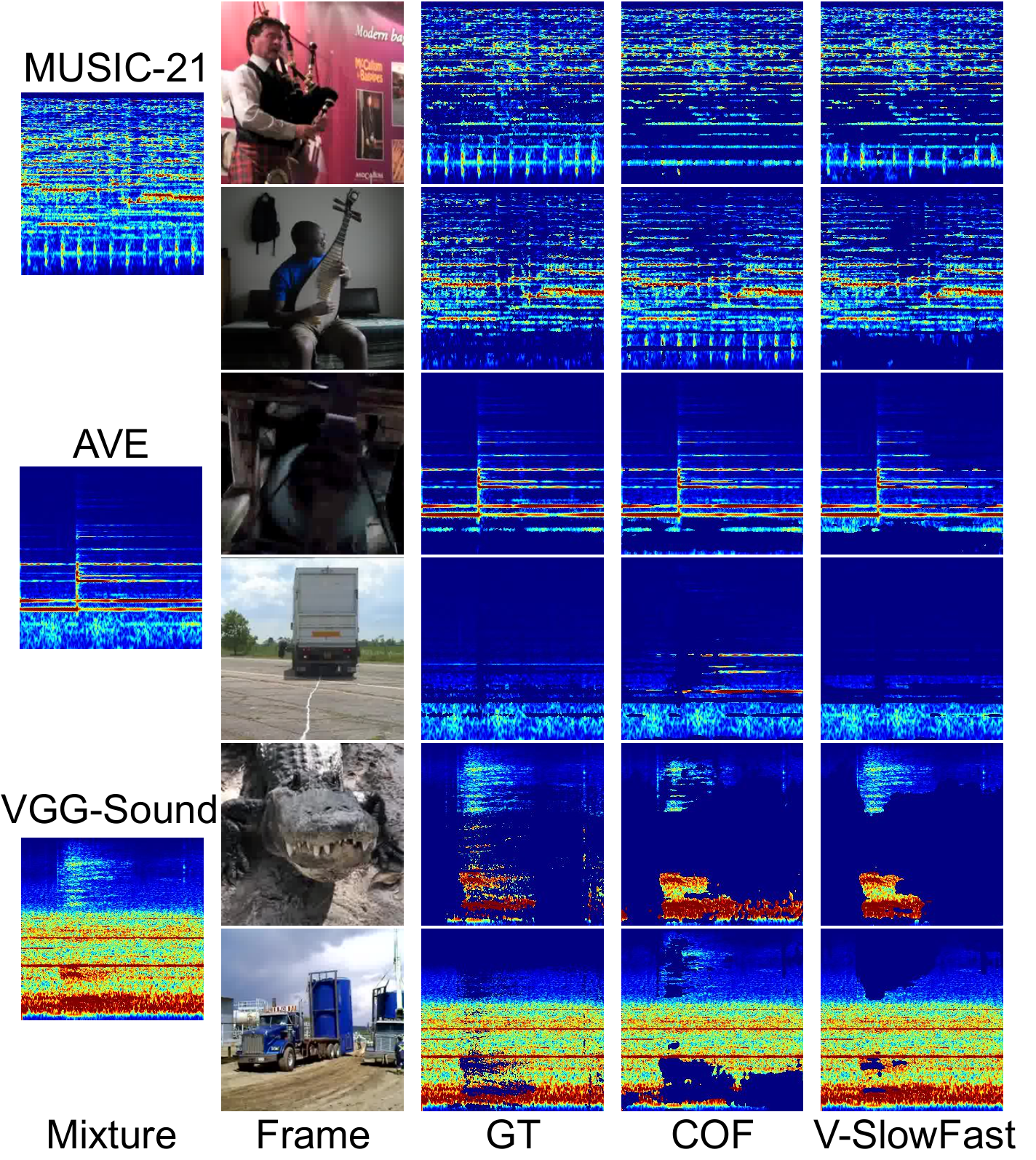}
    \end{tabular}
    \caption{Visualization of the source separation results with the MUSIC-21, AVE, and VGG-Sound datasets.}
    \label{fig:vis_sep}
\end{figure}

\setlength{\tabcolsep}{2pt}
\begin{table*}[!tbp]
    \centering
    \begin{tabular}{lccc|ccc|ccc|cc}
        \hline
        \multirow{2}{*}{Models} &
            \multicolumn{3}{c}{MUSIC-21} &
            \multicolumn{3}{c}{AVE} &
            \multicolumn{3}{c}{VGG-Sound} &
        \multirow{2}{*}{Param (M)} &
        \multirow{2}{*}{GMACs}\\
        & SDR & SIR & SAR & SDR & SIR & SAR & SDR & SIR & SAR \\
        \hline
        Sound of Pixels~\cite{Zhao_2018_ECCV} & 7.52 & 13.01 & 11.53 & 1.21 & 7.08 & 6.84 & 1.40 & 8.56 & 7.80 & 34.83 & 1.62\\
        Minus-Plus~\cite{Xu_2019_ICCV} & 9.15 & 15.38 & 12.11 & 1.96 & 7.95 & 8.08 & 1.93 & 9.30 & 8.25 & 58.35 & 5.13\\
        Cascaded Opponent Filter (appearance)~\cite{zhu2020visually} & 9.50 & 16.12 & 12.30 & 2.68 & 8.18 & 8.48 & 1.93 & 9.36 & 8.36 & 58.38 & 4.52\\
        V-SlowFast (1) & \bf 10.89 & \bf 18.33 & \bf 12.97 & \bf 3.09 & \bf 9.69 & 8.45 & \bf 2.59 & \bf 9.99 & 8.69 & 32.22 & 2.17\\
        V-SlowFast (2) & 9.54 & 16.02 & 12.26 & 2.92 & 9.68 & \bf 10.08 & 2.11 & 8.15 & \bf 13.35 & \bf 15.07 & \bf 0.84\\
        \hline
    \end{tabular}
    \caption{Source separation performance using mixtures of two sources from the MUSIC-21, AVE and VGG-Sound datasets with the single-frame based models of Sound of Pixels~\cite{Zhao_2018_ECCV}, Minus-Plus~\cite{Xu_2019_ICCV}, Cascaded Opponent Filter~\cite{zhu2020visually} and V-SlowFast. V-SlowFast (1) represents the model using V: MV2$^{+}$, Slow: DeepLabV3Plus and Fast: U-Net (9-layer). V-SlowFast (2) indicates the model of V: MV2$^{+}$, Slow: DeepLabV3Plus and Fast: DeepLabV3Plus. MV2$^{+}$ represents the vision network of \textit{MV2 + AVGA + Contrast}. The V-SlowFast results of $\alpha_{s}$=2, and $\alpha_{f}$=1 are reported, and the overall best model results are \textbf{bolded}.}
    \label{tab:exp_sota}
\end{table*}
\setlength{\tabcolsep}{2pt}

\paragraph{\bf Decoder}
The global visual-attended sound features are passed to the up-convolutional decoder ($\mathit{Slow^{D}}$) to produce a spectrum mask $\mathit{f_{Slow^{D,V}, n, \alpha_{s}}} \in \mathit{\mathbb{R}^{1 \times H_{S} \times W_{S1}}}$, which has the same-size as the temporally downsampled input spectrogram $\phi(X_{mix}, \alpha_{s})$. In order to make the output to have the same resolution as the original spectrogram, we apply $\phi^{inv}$ (inverse $\phi$) operation with the stride value of $\alpha_{s}$. Following sigmoid operation results in a binary mask $\mathit{\hat{B}_{Slow,n,1}}$, which is multiplied with the original input mixture spectrogram to produce an estimate of the component audio corresponding to the visual input. More formally,
\begin{equation}
    \centering
    \begin{split}
        \mathit{f_{Slow^{D,V}, n, \alpha_{s}}} = &\mathit{Slow^{D}\big(f_{Slow^{E,V}, n, \alpha_{s}}\big)}, \\
        \mathit{f_{Slow^{D,V}, n, 1}} = &\mathit{\phi^{inv}(f_{Slow^{D,V}, n, \alpha_{s}}, \alpha_{s})}, \\
        \mathit{\hat{B}_{Slow,n,1}} = &\mathit{\sigma (f_{Slow^{D,V}, n, 1})}, \\
        \mathit{\hat{X}_{Slow,n,1}} = &\mathit{\hat{B}_{Slow,n,1} \otimes X_{mix}},
    \end{split}
    \label{eq:ss_decoder}    
\end{equation}
where the $\phi^{inv}$ represents the inverse $\phi$ operation. $\mathit{f_{Slow^{D,V}, n, 1}}$ indicates the separated sound features for the $n$-th source ($n$-th input frame) from the slow spectrogram network, which has the full temporal resolution. $\sigma$ represents the sigmoid operation. $\otimes$ denotes the element-wise product. The output spectrogram $\mathit{\hat{X}_{Slow,n,1}}$ is formulated by element-wise multiplying the binary mask $\mathit{\hat{B}_{Slow,n,1}}$ with the original mixture spectrogram $X_{mix}$.

\subsection{Fast Spectrogram Residual Network}
\label{sec:fast_residual}

The fast spectrogram residual network also uses an encoder-decoder structure. The coarsely separated spectrogram from the previous slow spectrogram network and the original mixture spectrogram are concatenated first then forwarded to the $\phi$ operation (with $\alpha_{f}$, $\alpha_{f} < \alpha_{s}$). With the global attention from the visual embedding  $e_{v,n}$, the fast spectrogram residual network produces a residual spectrum mask $\mathit{f_{Fast^{D,V},n,1}}$, which is added to the spectrum mask $\mathit{f_{Slow^{D,V},n,1}}$ (from the previous slow spectrogram network). The final output is formed by multiplying the original mixture spectrogram with a binary mask obtained from the fast spectrogram residual network. More specifically, 
\begin{equation}
    \centering
    \begin{split}
        \mathit{f_{Fast^{E}, n, \alpha_{f}}} = & \mathit{Fast^{E}\big(\phi(cat[X_{mix}, \hat{X}_{Slow,n,1}], \alpha_{f})\big)}, \\
        \mathit{f_{Fast^{E,V}, n, \alpha_{f}}} = & \mathit{e^{T}_{v,n} \odot e_{v,n} \odot f_{Fast^{E}, n, \alpha_{f}}}, \\
        \mathit{f_{Fast^{D,V}, n, \alpha_{f}}} = & \mathit{Fast^{D}\big(f_{Fast^{E,V}, n, \alpha_{f}}\big)}, \\
        \mathit{f_{Fast^{D,V}, n, 1}} = & \mathit{\phi^{inv}(f_{Fast^{D,V}, n, \alpha_{f}}, \alpha_{f})}, \\
        \mathit{\hat{B}_{Fast,n,1}} = &\mathit{\sigma (f_{Slow^{D,V},n,1} \oplus f_{Fast^{D,V},n,1})}, \\
        \mathit{\hat{X}_{Fast,n,1}} = &\mathit{\hat{B}_{Fast,n,1}} \odot \mathit{X_{mix}}
    \end{split}
    \label{eq:fast}    
\end{equation}


\subsection{Overall Learning Objective}
\label{sec:objective}

The entire system is trained using a self-supervised setup with a large set of unlabelled videos. We formulate the visual sound separation learning objective to estimate the binary mask $\hat{B}_{n}$ to obtain the final output spectrogram (Eq.~\ref{eq:ss_encoder}, \ref{eq:avga}, \ref{eq:ss_decoder}, and~\ref{eq:fast}). The ground truth mask $B_{n}$ is formed as, 
\begin{equation}
    \centering
    B_{n}(f,t)= \llbracket X_{n}(f,t) \geq X_{m}(f,t)\rrbracket
\end{equation}
where $(f,t)$ represents the frequency-time coordinates in the sound spectrogram $X$. $N$ is the number of sources in the mixture and $\forall m \in (1, \dots, N)$. The \textbf{V-SlowFast} network is optimized by minimizing the binary cross entropy (BCE) loss between the estimated binary masks $\hat{B}_{n}$ and the ground-truth binary masks $B_{n}$,
\begin{equation}
    \centering
    \mathcal{L}_{sep} =\sum^N_{n=1} \textit{BCE}(\hat{B}_{Slow,n,1}, B_{n}) + \textit{BCE}(\hat{B}_{Fast,n,1}, B_{n})
    \label{eq:objective_sep}    
\end{equation}
where $\hat{B}_{Slow,n,1}$ and $\hat{B}_{Fast,n,1}$ represent the predicted binary masks at slow spectrogram network and fast sectrogram residual network, respectively. Then the predicted mask is multiplied with the input full resolution mixture spectrogram to get a predicted sound spectrogram. Finally, we apply an inverse Short-time Fourier Transform (iSTFT) on the predicted spectrogram to reconstruct the waveform of separated sound. The overall learning objective is formed by combining the contrastive learning objective with the sound separation objective as $\mathcal{L} = \mathcal{L}_{contrast} + \mathcal{L}_{sep}$.

\setlength{\tabcolsep}{2pt}
\begin{table}[!tbp]
    \centering
    \begin{tabular}{lccc|ccc}
        \hline
        \multirow{2}{*}{Models $\backslash$ N} &
            \multicolumn{3}{c}{N=3} &
            \multicolumn{3}{c}{N=4}\\
        & SDR & SIR & SAR \quad & SDR & SIR & SAR \\
        \hline
        Sound of Pixels~\cite{Zhao_2018_ECCV}  & 2.31 & 9.34 & 5.77 & -0.22 & 6.99 & 3.80 \\
        Minus-Plus~\cite{Xu_2019_ICCV} & 3.36 & 9.22 & 7.15 & 0.95 & 6.88 & 4.95 \\
        COF~\cite{zhu2020visually} & 4.08 & 9.95 & 7.68 & 0.97 & 7.19 & 5.05 \\
        V-SlowFast (1) & \bf 5.57 & \bf 12.59 & \bf 8.17 & \bf 2.20 & \bf 10.02 & \bf 5.11 \\
        V-SlowFast (2) & 4.16 & 11.91 & 6.94 & 1.08 & 8.49 & 4.47 \\
        \hline
    \end{tabular}
    \caption{Source separation performance using mixtures of three and four sources from the MUSIC-21 dataset using Sound of Pixels~\cite{Zhao_2018_ECCV}, Minus-Plus~\cite{Xu_2019_ICCV}, COF~\cite{zhu2020visually} and V-SlowFast.}
    \label{tab:more_sources}
\end{table}
\setlength{\tabcolsep}{2pt}


\section{Experiments}
\label{sec:exp}

In this section, we evaluate the visual sound separation performance of the proposed model.

\subsection{Datasets and Implementation details}
\label{sec:dataset}


We train and evaluate the proposed methods using small- and large-scale datasets: MUSIC-21~\cite{Zhao_2019_ICCV}, AVE~\cite{tian2018audio} and VGG-Sound~\cite{chen2020vggsound}. MUSIC-21~\cite{Zhao_2019_ICCV} contains 1365 videos from 21 instrumental categories. The AVE~\cite{tian2018audio} dataset, a subset of AudioSet~\cite{gemmeke2017audio}, contains 4143 10-second videos covering 28 audio-visual event categories. VGG-Sound~\cite{chen2020vggsound} is a recently released large-scale dataset with over 200k video clips for 310 categories of general classes. 




We follow the same setup as in~\cite{gan2020music,zhu2021visually}. The datasets are split into disjoint train, val (not used for MUSIC-21), and test sets. The audio mixtures are obtained by adding the audio tracks from N videos (N depends on test setup). We apply temporal downsampling operation $\phi$ on spectrograms with a downsampling rate $\alpha$ ($\textbf{Slow}$: $\alpha_{s}$ and $\textbf{Fast}$: $\alpha_{f}$). For instance, given a spectorgram with full temporal resolution of $T$, $\alpha_{s}$=2 represents a slow spectrogram with temporal resolution of $T/\alpha_{s}$. The sound separation performance is measured in terms of: Signal to Distortion Ratio (SDR), Signal to Interference Ratio (SIR), and Signal to Artifact Ratio (SAR). For the measures of SDR and SIR, higher value indicates better performance (more details are presented in the supplementary material). 

\begin{figure}[t]
    \centering
    \begin{tabular}{c}
        \includegraphics[width=0.90\linewidth]{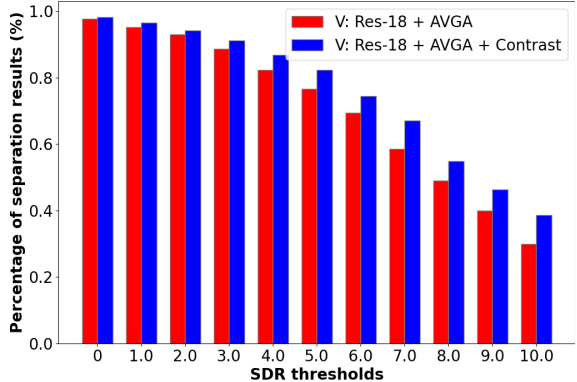}
    \end{tabular}
    \caption{Percentage of separation results over SDR thresholds.}
    \label{fig:V_barchart}
\end{figure}

\setlength{\tabcolsep}{2pt}
\begin{table}[!tbp]
    \centering
    \begin{tabular}{lc}
        \hline
        Models & SDR \\
        \hline
        Copy-Paste~\cite{zhu2021visually} & 4.39 \\
        Sound of Pixels~\cite{Zhao_2018_ECCV}  & 6.23 \\
        Minus-Plus~\cite{Xu_2019_ICCV} & 7.11 \\
        V-SlowFast (1) & \bf 8.64 \\
        V-SlowFast (2) & 8.51 \\
        \hline
    \end{tabular}
    \caption{Separating sounds in MUSIC-21 from background noises.}
    \label{tab:bg}
\end{table}
\setlength{\tabcolsep}{2pt}

\subsection{Source Separation with V-SlowFast Network}
\label{sec:exp_slow_fast}



\paragraph{\bf Separating Two Sound Sources}

Table~\ref{tab:exp_sota} summarizes the results in comparison with recent single frame methods Sound of Pixels~\cite{Zhao_2018_ECCV}, Minus-Plus~\cite{Xu_2019_ICCV} and Cascaded Opponent Filter (COF)~\cite{zhu2020visually} on MUSIC-21, AVE and VGG-Sound datasets using mixtures of two sound sources (N=2). We observe that our method consistently outperforms all baselines. Impressively, our system V-SlowFast (1) outperforms previous state-of-the-art multi-stage method~\cite{zhu2020visually} by 1.39dB on MUSIC-21, 0.41dB on AVE, and 0.66dB on VGG-Sound in terms of SDR while having substantially less parameters and small computational cost. Figure~\ref{fig:vis_sep} illustrates qualitative examples and additional examples are provided in the supplementary material. V-SlowFast (2) can achieve similar result as multi-stage approach~\cite{zhu2020visually} and better performance than recursive model~\cite{Xu_2019_ICCV} while using 74.2\% less parameters and 81.4\% less operations. These quantitative and qualitative results suggest that our model successfully exploits the explicit slow and fast spectrogram separately to improve the sound separation quality and to substantially reduce the total model size and computation.



\setlength{\tabcolsep}{1pt}
\begin{table}
    \centering
    \begin{tabular}{lcccccccc}
        \hline
        $\alpha$ & V & AVGA & Contrast & SDR & SIR & SAR & Param (M) & GMACs\\
        \hline
        \multirow{3}{*}{$\alpha$=1} &
        \checkmark & \ding{55} & \ding{55} & 8.06 & 14.79 & 10.82 & 36.03 & 2.04 \\
        & \checkmark & \checkmark & \ding{55} & 8.11 & 14.96 & 10.91 & 31.48 & 1.75\\
        & \checkmark & \checkmark & \checkmark & 8.69 & 15.72 & 11.02 & 31.48 & 1.75\\
        \hline
        \multirow{3}{*}{$\alpha$=2} &
        \checkmark & \ding{55} & \ding{55} & 7.78 & 13.70 & 11.09 & 36.03 & 1.19\\
        & \checkmark & \checkmark & \ding{55} & 8.02 & 14.46 & 11.11 & 
        31.48 & 1.04\\
        & \checkmark & \checkmark & \checkmark & 8.61 & 14.90 & 11.35 & 31.48 & 1.04\\
        \hline
        \multirow{3}{*}{$\alpha$=4} &
        \checkmark & \ding{55} & \ding{55} & 7.03 & 12.30 & 11.05 & 36.03 & 0.77\\
        & \checkmark & \checkmark & \ding{55} & 7.37 & 13.38 & 10.81 & 31.48 & 0.69\\
        & \checkmark & \checkmark & \checkmark & 7.93 & 13.96 & 11.02 & 31.48 & 0.69\\
        \hline
        \multirow{3}{*}{$\alpha$=8} &
        \checkmark & \ding{55} & \ding{55} & 6.08 & 10.85 & 11.04 & 36.03 & 0.55\\
        & \checkmark & \checkmark & \ding{55} & 6.43 & 11.11 & 11.17 & 31.48 & 0.52\\
        & \checkmark & \checkmark & \checkmark & 6.93 & 11.83 & 11.43 & 31.48 & 0.52\\
        \hline
        \multirow{3}{*}{$\alpha$=16} &
        \checkmark & \ding{55} & \ding{55} & 4.88 & 8.85 & 11.33 & 36.03 & 0.45\\
        & \checkmark & \checkmark & \ding{55} & 5.07 & 9.54 & 11.06 & 31.48 & 0.43\\
        & \checkmark & \checkmark & \checkmark & 5.66 & 9.88 & 11.12 & 31.48 & 0.43\\
        \hline
    \end{tabular}
    \caption{Source separation performance using vision embeddings from vison network $V$ of \textit{Res-18},  \textit{Res-18 + AVGA}, and  \textit{Res-18 + AVGA + Contrast}, and sound features from \textit{U-Net (7-layer)} on mixtures of two sources from the MUSIC-21 dataset.}
    \label{tab:1_stage_V}
\end{table}
\setlength{\tabcolsep}{1pt}

\paragraph{\bf Separating More Sound Sources}
A more challenging task is to separate a sound mixture that contains more than two sources. To this end, we assess the approaches by separating mixtures of three and four sources using MUSIC-21 dataset. We report the separation performance of V-SlowFast and the baselines~\cite{Zhao_2018_ECCV,Xu_2019_ICCV,zhu2020visually} in Table~\ref{tab:more_sources}. V-SlowFast outperforms the baselines with a clear margin for separating mixtures of three and fours sources. Qualitative examples are provided in the supplementary material.



\paragraph{\bf Separating Sound from Background Noises}
Due to the lack of ground truth for the source, assessing the performance in fully natural scenarios is difficult. However, we collect 100 natural background audios (retrieved from YouTube with keyword ``background noise'') to mix with available sources. Table~\ref{tab:bg} shows that our V-SlowFast models outperform all baselines on separating target sound from noisy mixture. ``Copy-Paste''~\cite{zhu2021visually} uses input mixture as output. Note that the previous work of Cascaded Opponent Filter~\cite{zhu2020visually} requires the knowledge of all the presenting sources within the sound mixture to separate each sound source. Instead of relaying on the visual cues of other sources, our V-SlowFast model is proposed to efficiently separate the interested sound with only its associated visual information. 

\subsection{Ablation Study}
\label{sec:ablation}



\paragraph{\bf Different Spectrogram Resolutions}


In Table~\ref{tab:1_stage_V}, we report the visual sound separation performance on different spectrogram temporal resolutions ($\alpha \in$ \{1, 2, 4, 8, 16\}). Similar as~\cite{Zhao_2018_ECCV}\footnote{Differently, the spatial size of the U-Net encoder output is 16 times smaller than input spectrogram in our methods instead of 128 times in~\cite{Zhao_2018_ECCV}.}, the separation mask is obtained by a linear multiplication between the visual embedding (vision network: Res-18) and the sound features (sound spectrogram network: 7-layer U-Net). We observe that with a larger value of downsampling rate $\alpha$ (lower temporal resolution), the model converges earlier. The smaller temporal resolution the input spectrogram has, the lower evaluation scores of SDR and SIR the models obtain. More details are reported in the supplementary material.




\paragraph{\bf Audio-Visual Global Attention}
\label{sec:exp_AVGA}


We assess the experiment results of using AVGA module together with vison network of Res-18 and sound spectrogram network of U-Net (7-layer) in Table~\ref{tab:1_stage_V}. As we can see, adopting the AVGA module results in a smaller model and less computations while obtaining separation scores improvement for all the $\alpha$, e.g. SDR gain of 0.34dB ($\alpha$=4).

\begin{figure}[t]
    \centering
    \begin{tabular}{c}
        \includegraphics[width=0.95\linewidth]{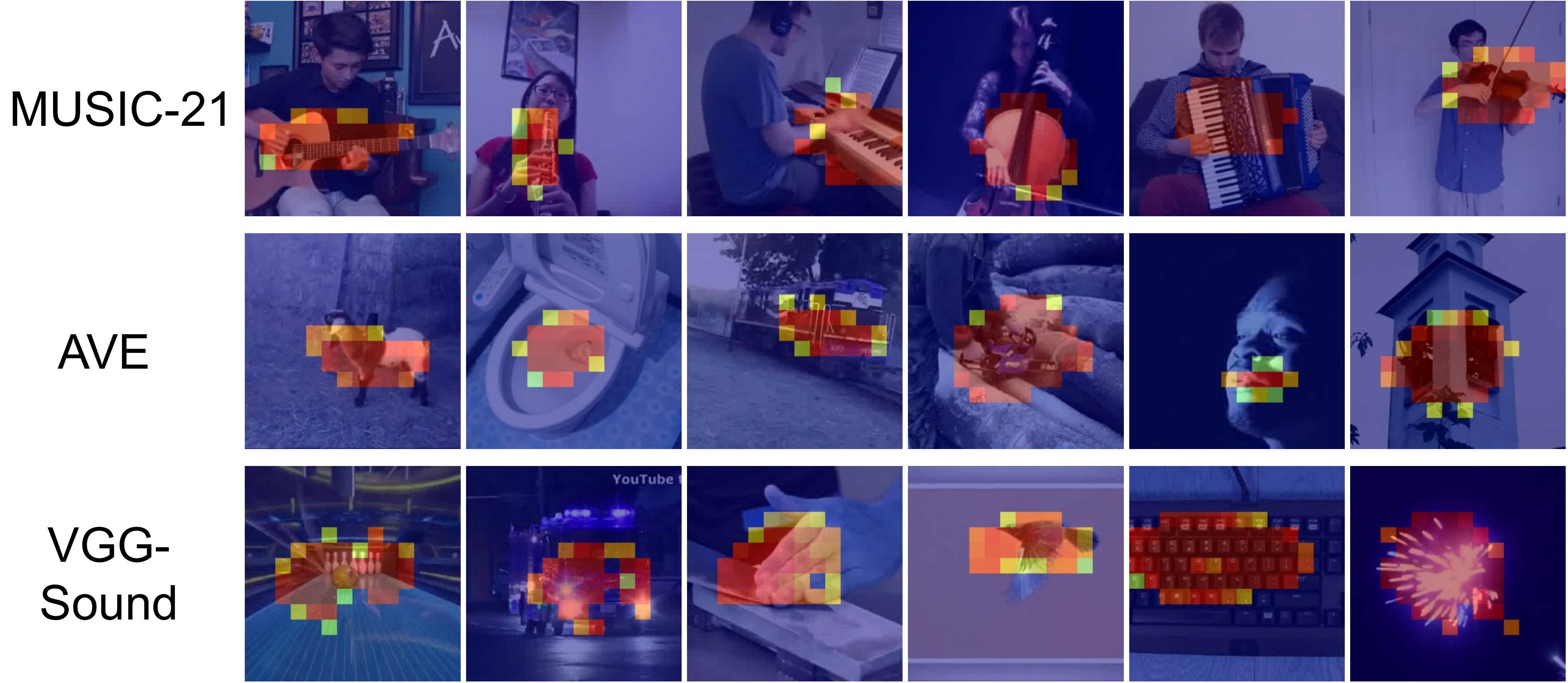}
    \end{tabular}
    \caption{Visualization of the sounding sources with the MUSIC-21, AVE, and VGG-Sound datasets.}
    \label{fig:vis_loc}
\end{figure}

\paragraph{\bf Visual Contrastive Learning}
\label{sec:exp_contrastAA}

We report the experiments result of using visual contrastive learning in Table~\ref{tab:1_stage_V} ($\mathit{Contrast}$). Note that the contrastive learning objective is only considered during training procedure. Thus, it does not bring extra operations for inference. As reported in Table~\ref{tab:1_stage_V}, the visual contrastive learning improves the separation score by the gain of around 0.6dB in SDR for all the $\alpha$. 


For better visualizing the improvement the visual contrastive learning brings to \textit{Res-18 + AVGA} model, we display the bar chart percentage of separation results over a wide range of SDR thresholds in Figure~\ref{fig:V_barchart}. The performance gap shows that the models using contrastive learning surpasss baseline with a large margin especially when the SDR threshold is $\geq$ 7.0. The contrastive learning allows the vision network to learn discriminative visual features and further improve the separation performance. 

During inference, the sounding source location mask $M_{loc}$ is yielded by an inner product between the visual feature and its own visual embedding (see Figure~\ref{fig:overview}). Examples are reported  in Figure~\ref{fig:vis_loc} and in the supplementary material.




\paragraph{\bf Architecture Variants of the Sound Network}
\label{sec:exp_model_variants}
The architecture of the above mentioned sound spectrogram network is 7 up- and 7 down-convolutional layers U-Net, which is referred as U-Net (7-layer). In this section, we study how the model performs on different spectrogram resolutions when using less (5-layer) or more (9-layer) convolutional layers U-Net as the sound spectrogram network. Table~\ref{tab:1_stage_SS_unet} summarises the evaluation metrics, number of parameters and operations when using vision network of \textit{Res-18 + AVGA + Contrast} and sound spectrogram network of U-Net (5-, 7-, 9-layer) with different $\alpha$. We observe performance decrease when using shallower U-Net, and performance increase when using deeper U-Net for all the $\alpha$. In addition, for smaller $\alpha$ (high temporal resolution), the performance increases a larger margin (e.g. 1.09dB in SDR of $\alpha$=1) when switching from U-Net (7-layer) to deeper U-Net (9-layer) in comparison with counterparts of larger $\alpha$ (e.g. 0.58dB in SDR of $\alpha$=16). Furthermore, using U-Net (9-layer) only increases total 2.22M parameters, and using U-Net (5-layer) outcomes a model with total 17.32M parameters.

\setlength{\tabcolsep}{2pt}
\begin{table}
    \centering
    \begin{tabular}{lcccccc}
        \hline
        $alpha$ & U-Net & SDR & SIR & SAR & Param (M) & GMACs\\
        \hline
        \multirow{3}{*}{$\alpha$=1} & 
        5-layer & 7.11 & 14.33 & 9.51 & 17.32 & 1.34 \\
        & 7-layer & 8.69 & 15.72 & 11.02 & 31.48 & 1.75\\
        & 9-layer & 9.78 & 17.13 & 12.00 & 33.70 & 2.15\\
        \hline
        \multirow{3}{*}{$\alpha$=2} & 
        5-layer & 6.96 & 13.62 & 9.99 & 17.32 & 0.84 \\
        & 7-layer & 8.61 & 14.90 & 11.35 & 31.48 & 1.04 \\
        & 9-layer & 9.55 & 16.09 & 12.10 & 33.70 & 1.25 \\
        \hline
        \multirow{3}{*}{$\alpha$=4} & 
        5-layer & 6.25 & 11.57 & 10.37 & 17.32 & 0.59 \\
        & 7-layer & 7.93 & 13.96 & 11.02 & 31.48 & 0.69 \\
        & 9-layer & 8.66 & 14.89 & 11.44 & 33.70 & 0.79 \\
        \hline
        \multirow{3}{*}{$\alpha$=8} & 
        5-layer & 5.71 & 10.54 & 10.83 & 17.32 & 0.47 \\
        & 7-layer & 6.93 & 11.83 & 11.43 & 31.48 & 0.52 \\
        & 9-layer & 7.60 & 12.75 & 11.82 & 33.70 & 0.57 \\
        \hline
        \multirow{3}{*}{$\alpha$=16} & 
        5-layer & 4.69 & 9.01 & 10.91 & 17.32 & 0.40 \\
        & 7-layer & 5.66 & 9.88 & 11.12 & 31.48 & 0.43 \\
        & 9-layer & 6.24 & 10.48 & 11.80 & 33.70 & 0.45 \\
        \hline
    \end{tabular}
    \caption{Source separation performance using sound network of 5, 7, and 9 up- and down-convolutional U-Net layers with vision network of  \textit{Res-18 + AVGA + Contrast} on mixtures of two sources from the MUSIC-21 dataset.}
    \label{tab:1_stage_SS_unet}
\end{table}
\setlength{\tabcolsep}{2pt}

\setlength{\tabcolsep}{4pt}
\begin{table*}[t]
    \centering
    \begin{tabular}{cccccccc}
        \hline
        V & Slow & Fast & SDR & SIR & SAR & Param (M) & GMACs\\
        \hline
        \multirow{4}{*}{Res-18$^{+}$} & 
        $\alpha_{s}$=2, U-Net (7-layer) & $\alpha_{f}$=1, U-Net (7-layer) & 10.43 & 17.88 & 12.62 & 51.69 & 2.45 \\
        & $\alpha_{s}$=4, U-Net (7-layer) & $\alpha_{f}$=1, U-Net (7-layer) & 10.33 & 17.37 & 12.62 & 51.69 & 2.10 \\
        & $\alpha_{s}$=8, U-Net (7-layer) & $\alpha_{f}$=1, U-Net (7-layer) & 9.69 & 16.84 & 12.12 & 51.69 & 1.93 \\
        & $\alpha_{s}$=16, U-Net (7-layer) & $\alpha_{f}$=1, U-Net (7-layer) & 9.47 & 16.50 & 11.87 & 51.69 & 1.84 \\
        \hline 
        \multirow{4}{*}{Res-18$^{+}$} &
        $\alpha_{s}$=2, U-Net (5-layer) & $\alpha_{f}$=1, U-Net (9-layer) & 10.55 & 18.10 & 12.54 & 39.74 & 2.65 \\
        & $\alpha_{s}$=4, U-Net (5-layer) & $\alpha_{f}$=1, U-Net (9-layer) & 10.36 & 17.87 & 12.42 & 39.74 & 2.40 \\
        & $\alpha_{s}$=8, U-Net (5-layer) & $\alpha_{f}$=1, U-Net (9-layer) & 9.94 & 17.27 & 12.30 & 39.74 & 2.28 \\
        & $\alpha_{s}$=16, U-Net (5-layer) & $\alpha_{f}$=1, U-Net (9-layer) & 9.92 & 17.37 & 12.22 & 39.74 & 2.22 \\
        \hline
        \multirow{5}{*}{Res-18$^{+}$} &
        $\alpha_{s}$=4, U-Net (5-layer) & $\alpha_{f}$=2, U-Net (9-layer) & 9.90 & 17.06 & 12.14 & 39.74 & 1.50 \\
        & $\alpha_{s}$=8, U-Net (5-layer) & $\alpha_{f}$=2, U-Net (9-layer) & 9.75 & 16.82 & 12.10 & 39.74 & 1.37 \\
        & $\alpha_{s}$=16, U-Net (5-layer) & $\alpha_{f}$=2, U-Net (9-layer) & 9.57 & 16.71 & 11.86 & 39.74 & 1.31 \\
        & $\alpha_{s}$=8, U-Net (5-layer) & $\alpha_{f}$=4, U-Net (9-layer) & 8.66 & 14.80 & 11.55 & 39.74 & 0.92 \\
        & $\alpha_{s}$=16, U-Net (5-layer) & $\alpha_{f}$=4, U-Net (9-layer) & 8.54 & 14.80 & 11.36 & 39.74 & 0.86 \\
        \hline
        \multirow{2}{*}{Res-18$^{+}$} &
        $\alpha_{s}$=2, DeepLabV3Plus & $\alpha_{f}$=1, U-Net (9-layer) & 10.98 & 18.27 & 12.95 & 38.97 & 2.39 \\
        & $\alpha_{s}$=4, DeepLabV3Plus & $\alpha_{f}$=2, U-Net (9-layer) & 10.38 & 17.29 & 12.59 & 38.97 & 1.37 \\
        \hline
        \multirow{2}{*}{MV2$^{+}$} &
        $\alpha_{s}$=2, DeepLabV3Plus & $\alpha_{f}$=1, U-Net (9-layer) & 10.89 & 18.33 & 12.97 & 32.22 & 2.17 \\
        & $\alpha_{s}$=4, DeepLabV3Plus & $\alpha_{f}$=2, U-Net (9-layer) & 10.39 & 17.25 & 12.69 & 32.22 & 1.15 \\
        \hline
        \multirow{2}{*}{MV2$^{+}$} &
        $\alpha_{s}$=2, DeepLabV3Plus & $\alpha_{f}$=1, DeepLabV3Plus & 9.54 & 16.02 & 12.26 & 15.07 & 0.84 \\
        & $\alpha_{s}$=4, DeepLabV3Plus & $\alpha_{f}$=2, DeepLabV3Plus & 8.64 & 14.89 & 11.54 & 15.07 & 0.48 \\
        \hline
    \end{tabular}
    \caption{Source separation performance of the V-SlowFast framework using mixtures of two sources from the MUSIC-21 dataset. Res-18$^{+}$ represents the vision network of  \textit{Res-18 + AVGA + Contrast} and MV2$^{+}$ represents the vision network of  \textit{MV2 + AVGA + Contrast}.}
    \label{tab:two-stage}
\end{table*}
\setlength{\tabcolsep}{4pt}

\paragraph{\bf Visually Guided SlowFast Sound Separation}

In this section, we examine the model performance when perceiving the slow and fast spectrograms separately. 

\textit{V-SlowFast:}
We firstly separate sources using the V-SlowFast network $\textbf{V}$: Res-18$^{+}$, $\textbf{Slow}$: U-Net (7-layer), and $\textbf{Fast}$: U-Net (7-layer) in Table~\ref{tab:two-stage} (with $\alpha_{s} \in \{2, 4, 8, 16\}$, $\alpha_{f}=1$). The results with both of the slow and fast spectrograms clearly surpass the network with only single spectrogram model in Table~\ref{tab:1_stage_V}, which proposes that treating the slow and fast spectrograms separately is important for the sound separation quality. Inspired by the observation from Table~\ref{tab:1_stage_SS_unet}, that U-Net (5-layer) is a very light model and U-Net (9-layer) can have large performance gain without bring heavy parameters and operations, we design the architecture of the V-SlowFast by using $\textbf{V}$: Res-18$^{+}$, $\textbf{Slow}$: U-Net (5-layer), and $\textbf{Fast}$: U-Net (9-layer). As is shown in Table~\ref{tab:two-stage}, with the compromise of around 0.2 $\sim$ 0.3 more GMACs, the method obtains performance gain (especially for the larger $\alpha_{s}$) while having 11.95M less parameters for all the experiemnts. 

In comparison, we also examine how the ``V-FastSlow'' performs, where the fast spectropgram appears first and slow spectrogram occurs second. The V-FastSlow results in similar performance as the V-SlowFast in terms of the evaluation metrics, number of parameters and operations (more details are presented in the supplementary material). Therefore, in this work, we discuss only on the case of V-SlowFast network.

%


\textit{Different combinations of $\alpha_{s}$ and $\alpha_{f}$:}
We further study how the V-SlowFast performs on different combinations of $\alpha_{s}$ and $\alpha_{f}$ using $\textbf{V}$: Res-18$^{+}$, $\textbf{Slow}$: U-Net (5-layer), and $\textbf{Fast}$: U-Net (9-layer) in Table~\ref{tab:two-stage}. We observe clear reduction in computations when using larger $\alpha_{f}$ (e.g. $\alpha_{f}$=2, 4). The system has a slight performance drop with the increasing of the $\alpha_{s}$, and the performance drops dramatically with the increasing of the $\alpha_{f}$, which suggest that the performance of the fast spectrogram network determines the overall result.


\textit{Smaller architecture variants:}
In order to separate sound sources more efficiently, we explore the system with smaller architecture variants, e.g. MV2~\cite{sandler2018mobilenetv2}. We adapt MV2 as the vision network (4.52M parameters and 0.12 GMACs), DeepLabV3Plus~\cite{chen2018encoder} (MV2 as backbone) as the slow and fast specgrogram networks (5.27M parameters). The performance of the combinations between different model variants for the V-SlowFast framework are presented in Table~\ref{tab:two-stage}. When using DeepLabV3Plus as the slow spectrogram network, and U-Net (9-layer) as the fast spectrogram residual network, the models with vision network of Res-18$^{+}$ and MV2$^{+}$ achieve similar results, e.g. Res-18$^{+}$: 10.98dB ($\alpha_{s}$=2, $\alpha_{f}$=1) and 10.38dB ($\alpha_{s}$=4, $\alpha_{f}$=2) of SDR in comparison with MV2$^{+}$: 10.89dB ($\alpha_{s}$=2, $\alpha_{f}$=1) and 10.39dB ($\alpha_{s}$=4, $\alpha_{f}$=2). Differently, the vision network of MV2$^{+}$ outcomes a model with 6.75M less parameters and 0.22 GMACs less operations. Thus, we refer the architecture of $\textbf{V}$: MV2$^{+}$, $\textbf{Slow}$: DeepLabV3Plus, and $\textbf{Fast}$: U-Net (9-layer) as $\textbf{V-SlowFast}$ (1). Furthermore, when adopting the DeepLabV3Plus as the architecture of the fast spectrogram residual network, the model $\textbf{V}$: MV2$^{+}$, $\textbf{Slow}$: DeepLabV3Plus, and $\textbf{Fast}$: DeepLabV3Plus obtains close results as the recent single frame based state-of-the-art method COF~\cite{zhu2020visually} while only contains total 15.07M model parameters and 0.84 GMACs, which is denoted as $\textbf{V-SlowFast}$ (2).





\section{Conclusions}

We proposed a new light yet efficient three-stream framework $\textbf{V-SlowFast}$ that operates on visual image, slow spectrogram, and fast spectrogram. We introduced two contrastive objectives to encourage the network to learn discriminative visual features for separating sounds and localizing sounding sources. In addition, we proposed an audio-visual global attention module for audio and visual features fusion. Furthermore, we studied
visually guided sound separation by treating the slow and fast spectrograms separately in terms of different temporal resolutions and model variants. The proposed $\textbf{V-SlowFast}$ models show excellent performance on small- and large-scale datasets: MUSIC-21, AVE, and VGG-Sound and can have 74.2\% reduction in number of parameters and 81.4\% reduction in GMACs compared to recent single-frame based baselines.


\paragraph{\bf Acknowledgement} This work is supported by the Academy of Finland (projects 327910 \& 324346).

{\small
\bibliographystyle{ieee_fullname}
\bibliography{main}
}


\vfill
\pagebreak

\appendix
\section*{Supplementary Material}


The supplementary material is arranged as follows: Section~\ref{sec:supp_alpha} reports the loss and matrix curves of the visual sound source separation on single spectrogram of different temporal resolutions; Section~\ref{sec:supp_fastslow} presents the visual sound separation performance with V-FastSlow framework; Section~\ref{sec:supp_vis} provides additional visualization of the source separation and localization; Section~\ref{sec:supp_implementation} contains additional implementation details.





\section{Visual Sound Separation on Single Spectrogram of Different Temporal Resolutions}
\label{sec:supp_alpha}

\begin{figure}[t]
    \centering
    \renewcommand\thefigure{A}  
    \begin{tabular}{c}
        \includegraphics[width=1.0\linewidth]{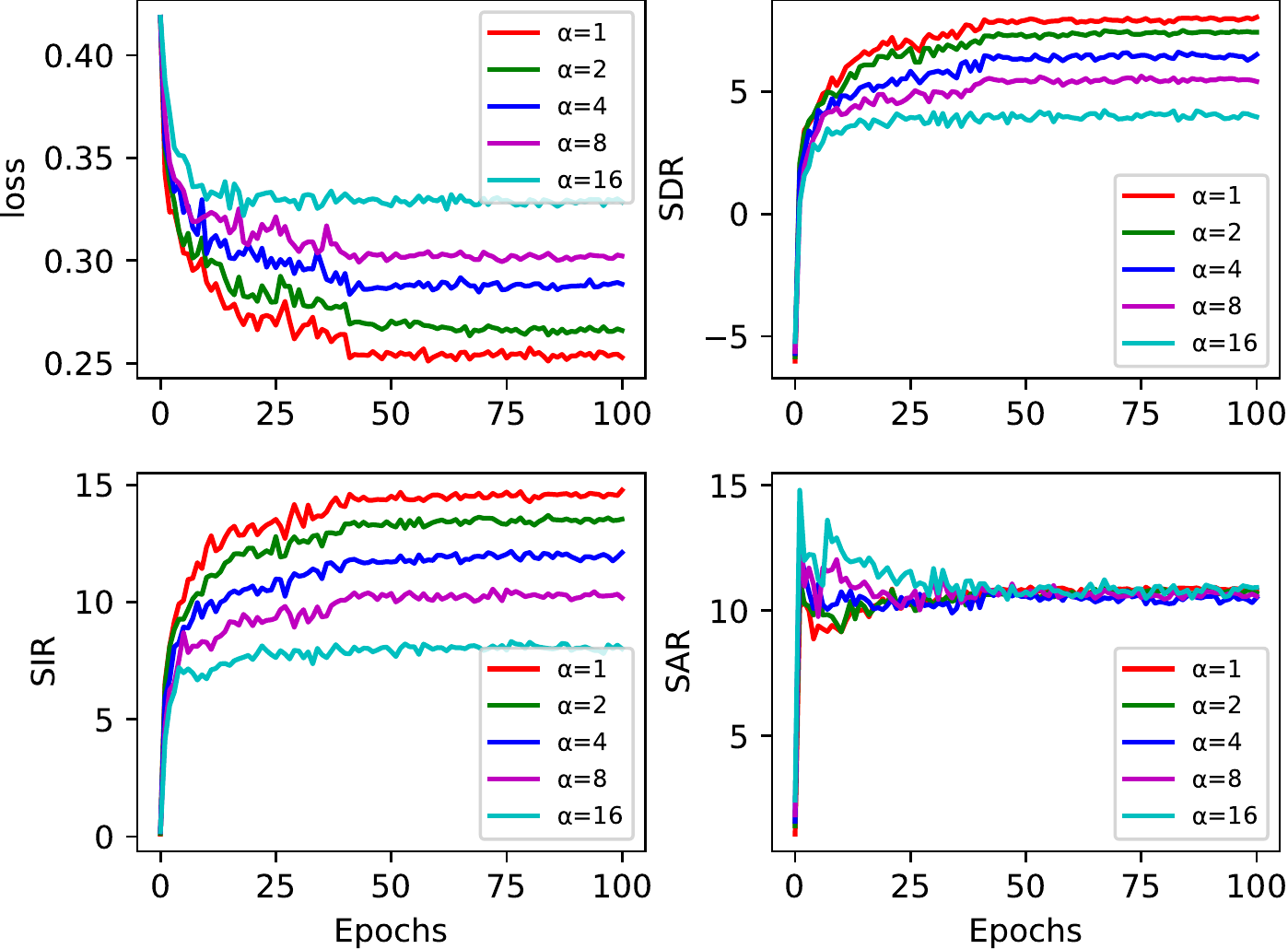}
    \end{tabular}
    \caption{Visualization of the loss and matrix curves of the visual sound source separation on single spectrogram of different temporal resolutions ($\alpha \in$ \{1, 2, 4, 8, 16\}).}
    \label{fig:supp_matrix}
\end{figure}

In Figure~\ref{fig:supp_matrix}, we display the loss and evaluation matrix curves (training procedure) of the visual sound source separation performance on single spectrogram of different temporal resolutions ($\alpha \in$ \{1, 2, 4, 8, 16\}). We observe that the training procedure with larger $\alpha$ converges faster. In addition, the smaller temporal resolution (larger $\alpha$) the input spectrogram has, the lower evaluation scores of SDR and SIR the models obtain, which reflects the larger separation loss. However, as is shown in Figure~\ref{fig:supp_matrix}, the SAR score does not follow the same trend. SAR captures only the absence of artifacts, hence can be high even if separation is poor. Thus, we conclude that the SDR and SIR scores measure the separation quality. 


\section{Visual Sound Separation with V-FastSlow Network}
\label{sec:supp_fastslow}

In order to study whether the order of the slow and fast spectrogram matters, we also assess the opposite way (V-FastSlow), where the fast spectropgram appears first and slow spectrogram occurs second. We experimented the V-FastSlow network with $\alpha_{f}=1$, $\alpha_{s} \in \{2, 4, 8, 16\}$ and reported the results in Table~\ref{tab:supp_v_fastslow}. The V-FastSlow obtains very close performance as the V-SlowFast in terms of the evaluation metrics, number of parameters and operations. Especially when the $\alpha_{s}$=2 and 4, the V-SlowFast achieves slightly better performance, e.g. gain of 0.2 $\sim$ 0.4dB in SDR. Thus, we mainly discuss on the case of V-SlowFast model in the main paper.

\begin{table*}[t]
    \centering
    \renewcommand\thetable{A} 
    \begin{tabular}{llccccc}
        \hline
        Fast & Slow & SDR & SIR & SAR & Param (M) & GMACs\\
        \hline
        $\alpha_{f}$=1, U-Net (7-layer) & $\alpha_{s}$=2, U-Net (7-layer) & 10.26 & 16.92 & 12.87 & 51.69 & 2.45 \\
        $\alpha_{f}$=1, U-Net (7-layer) & $\alpha_{s}$=4, U-Net (7-layer) & 10.11 & 16.97 & 12.59 & 51.69 & 2.10 \\
        $\alpha_{f}$=1, U-Net (7-layer) & $\alpha_{s}$=8, U-Net (7-layer) & 9.94 & 16.83 & 12.34 & 51.69 & 1.92 \\
        $\alpha_{f}$=1, U-Net (7-layer) & $\alpha_{s}$=16, U-Net (7-layer) & 9.58 & 16.62 & 11.92 & 51.69 & 1.84 \\
        \hline
        $\alpha_{f}$=1, U-Net (9-layer) & $\alpha_{s}$=2, U-Net (5-layer) & 10.11 & 16.50 & 12.73 & 39.74 & 2.65 \\
        $\alpha_{f}$=1, U-Net (9-layer) & $\alpha_{s}$=4, U-Net (5-layer) & 10.04 & 17.00 & 12.50 & 39.74 & 2.40 \\
        $\alpha_{f}$=1, U-Net (9-layer) & $\alpha_{s}$=8, U-Net (5-layer) & 9.96 & 17.06 & 12.35 & 39.74 & 2.28 \\
        $\alpha_{f}$=1, U-Net (9-layer) & $\alpha_{s}$=16, U-Net (5-layer) & 9.71 & 17.00 & 12.01 & 39.74 & 2.21 \\
        \hline
    \end{tabular}
    \caption{Source separation performance using mixtures of two sources from the MUSIC-21 dataset with V-FastSlow network for $\alpha_{f}$=1, $\alpha_{s} \in$ \{2, 4, 8, 16\}. The vision network is $\textit{Res-18 + AVGA + Contrast}$.}
    \label{tab:supp_v_fastslow}
\end{table*}
\section{Additional Qualitative Results}
This section provides additional qualitative visualization of the visual sound source separation and source localization examples. 
\label{sec:supp_vis}

\subsection{Visual Sound Separation}
\begin{figure*}[!thp]
    \centering
    \renewcommand\thefigure{B} 
    \includegraphics[width=1.0\linewidth]{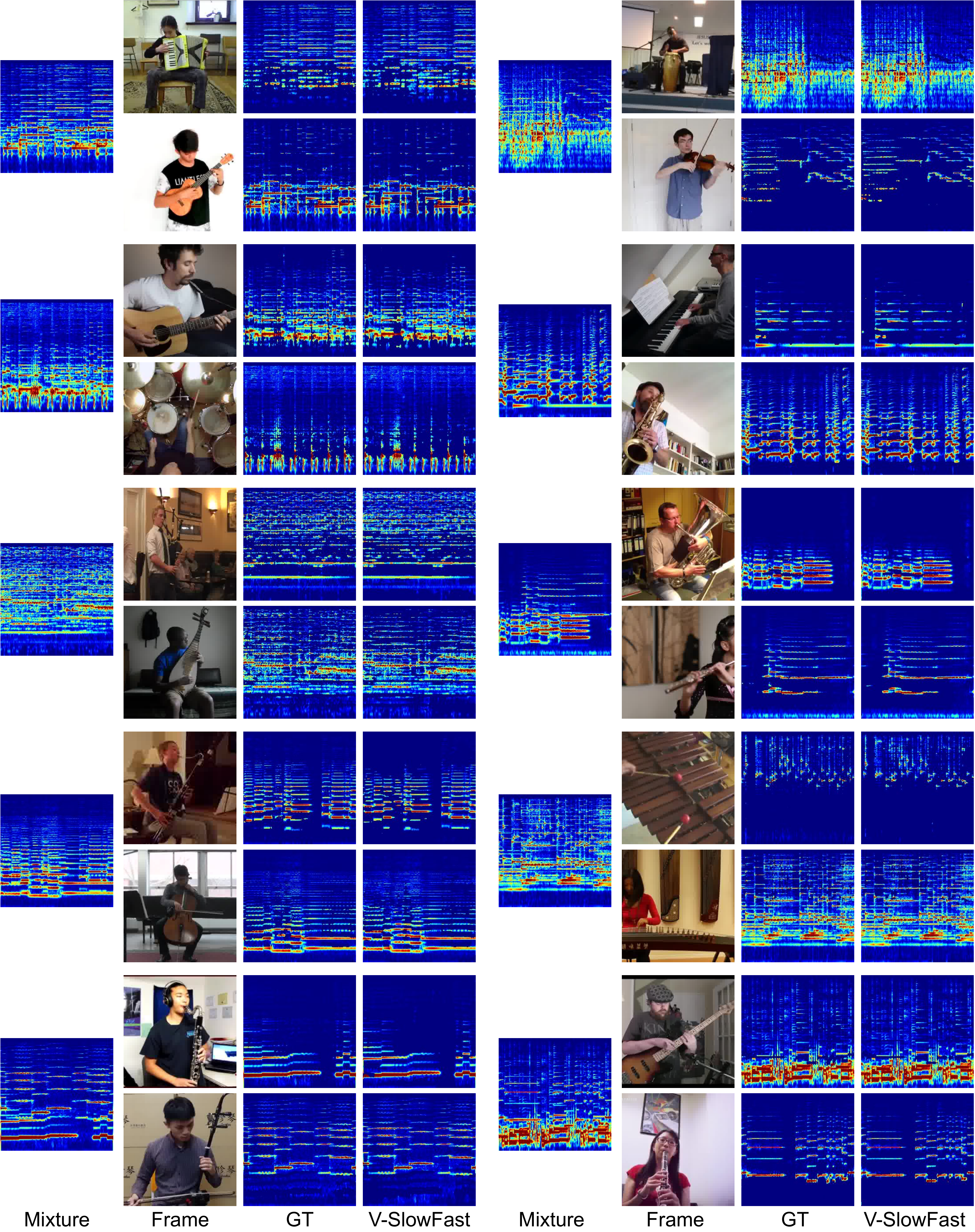}
   \caption{Visualization of the source separation results using V-SlowFast network with mixtures of two sources from MUSIC-21 dataset.}
\label{fig:vis_sep_MUSIC21_2}
\end{figure*}

\begin{figure*}[!thp]
    \centering
    \renewcommand\thefigure{C} 
    \includegraphics[width=1.0\linewidth]{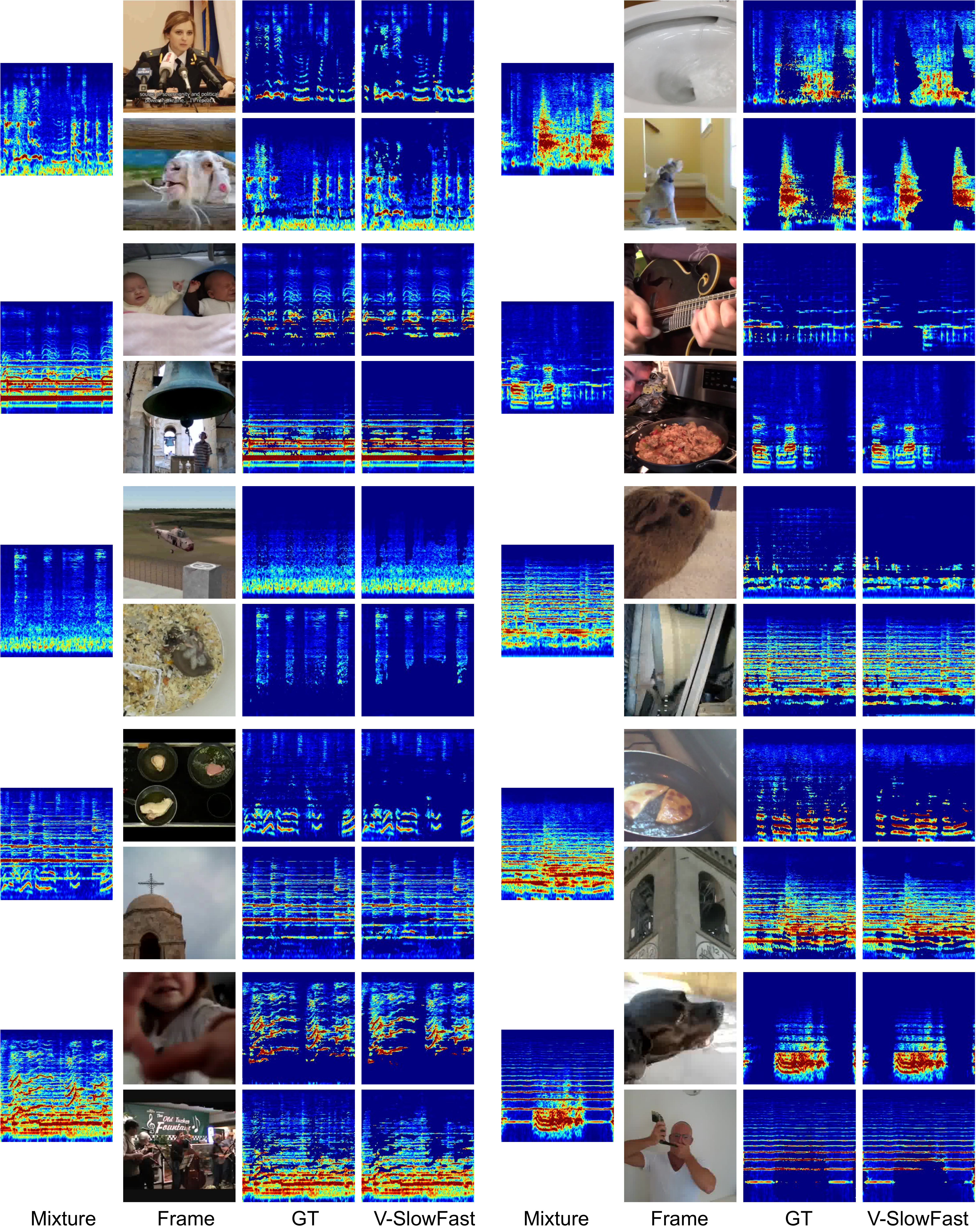}
   \caption{Visualization of the source separation results using V-SlowFast network with mixtures of two sources from AVE dataset.}
\label{fig:vis_sep_AVE_2}
\end{figure*}

\begin{figure*}[!thp]
    \centering
    \renewcommand\thefigure{D} 
    \includegraphics[width=1.0\linewidth]{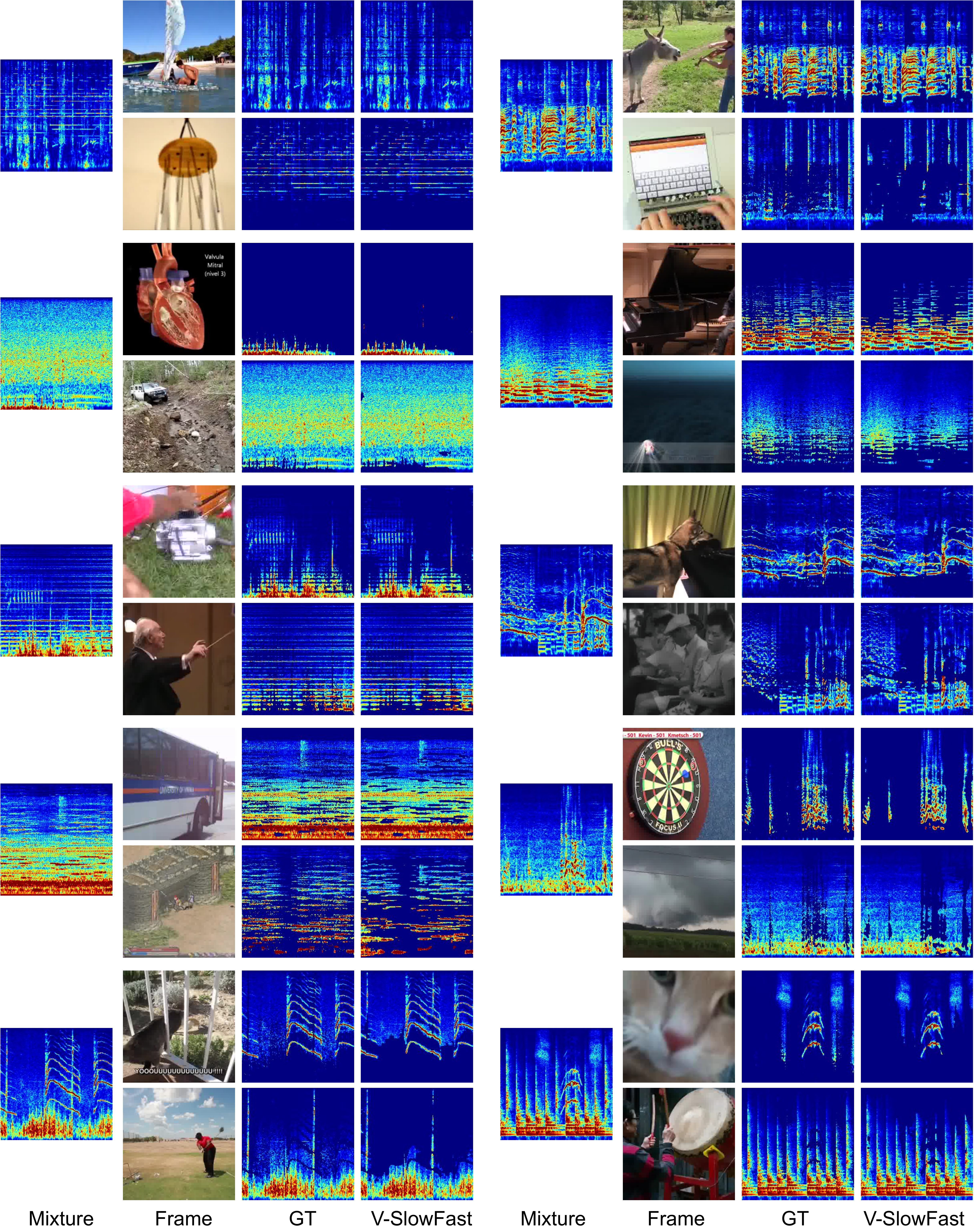}
   \caption{Visualization of the source separation results using V-SlowFast network with mixtures of two sources from VGG-Sound dataset.}
\label{fig:vis_sep_vggsound_2}
\end{figure*}

\begin{figure*}[!thp]
    \centering
    \renewcommand\thefigure{E} 
    \includegraphics[width=1.0\linewidth]{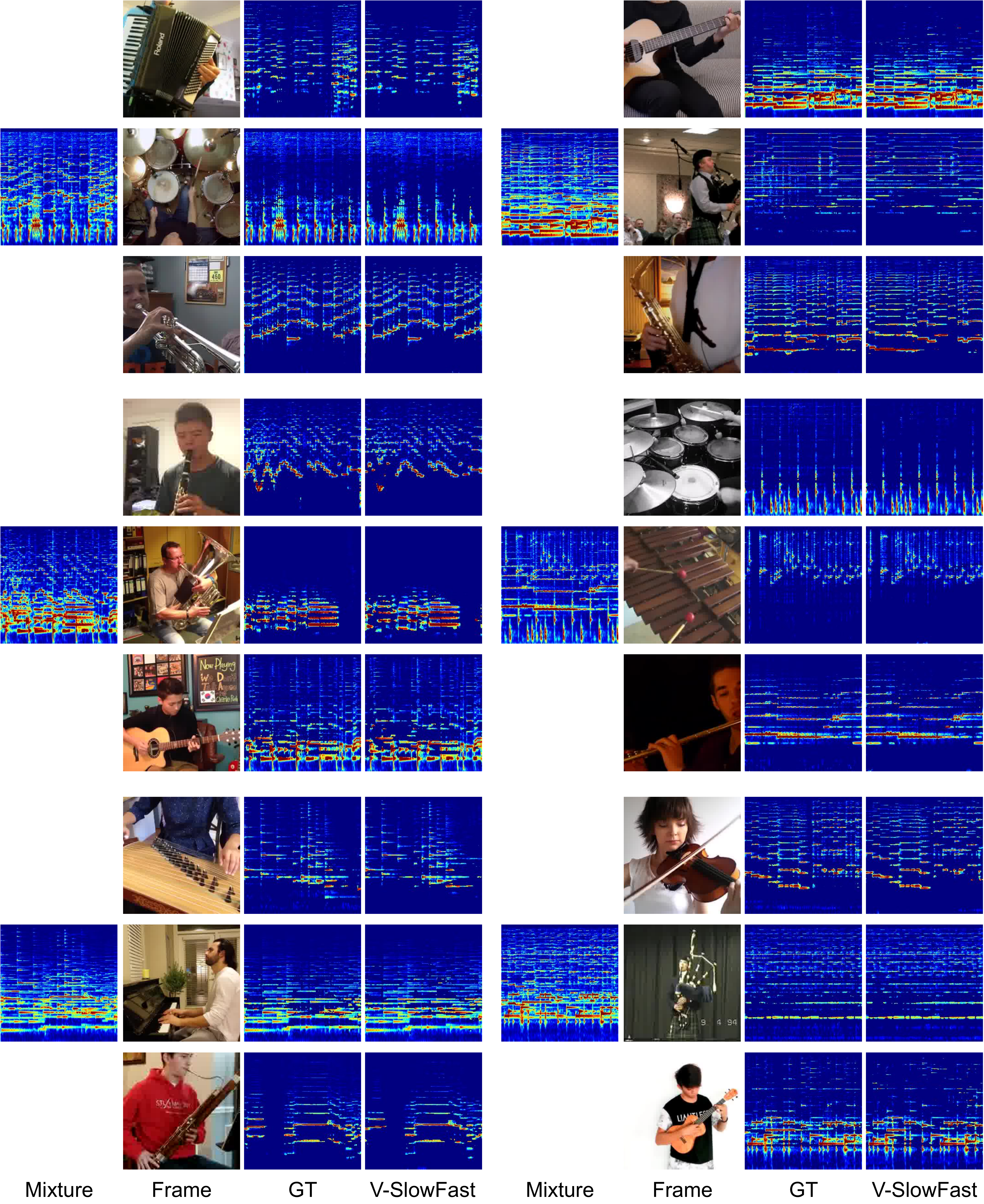}
   \caption{Visualization of the source separation results using V-SlowFast network with mixtures of three sources from MUSIC-21 dataset.}
\label{fig:vis_sep_MUSIC21_3}
\end{figure*}

\begin{figure*}[!thp]
    \centering
    \renewcommand\thefigure{F} 
    \includegraphics[width=1.0\linewidth]{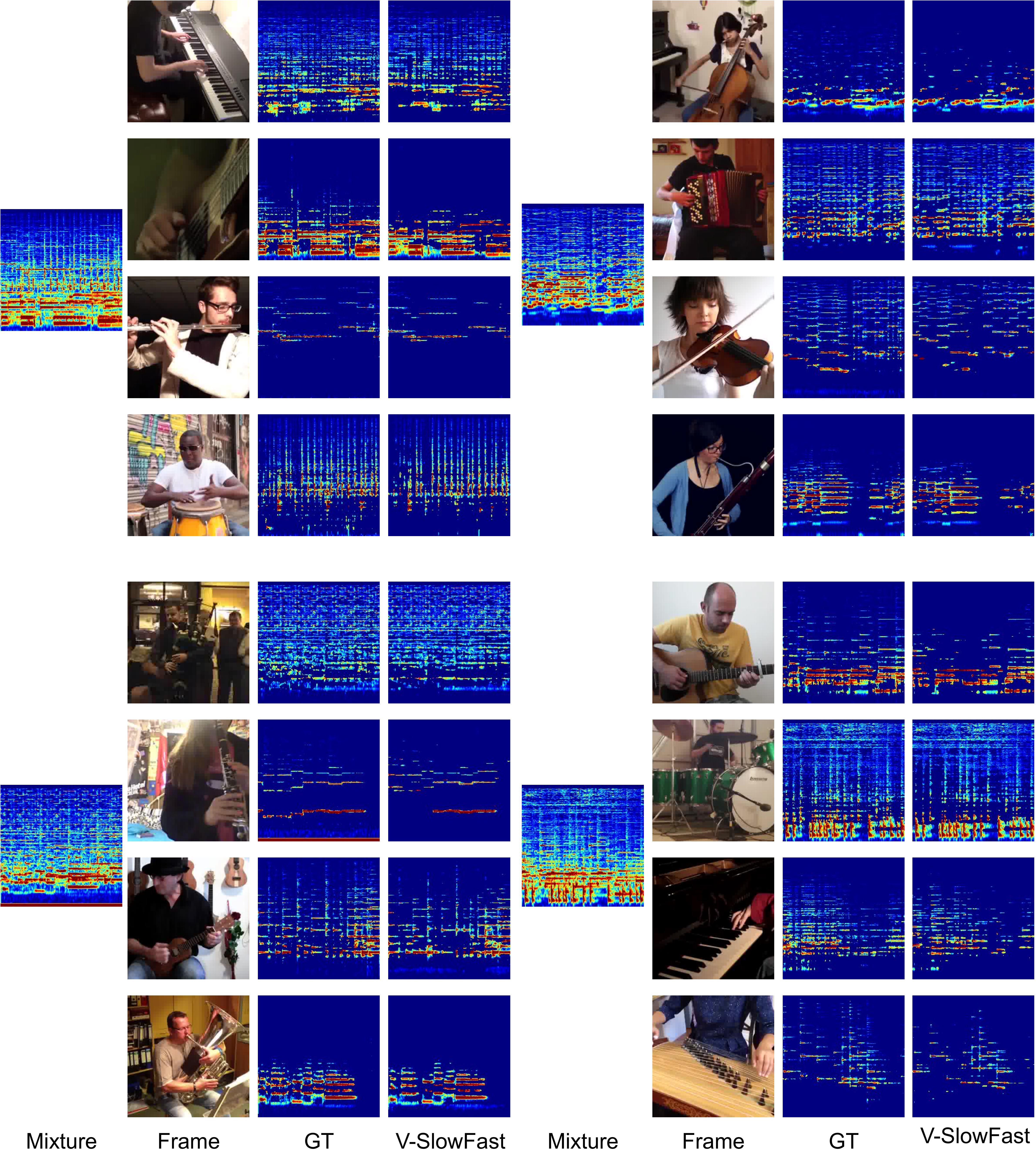}
   \caption{Visualization of the source separation results using V-SlowFast network with mixtures of four sources from MUSIC-21 dataset.}
\label{fig:vis_sep_MUSIC21_4}
\end{figure*}

Figures~\ref{fig:vis_sep_MUSIC21_2}, \ref{fig:vis_sep_AVE_2} and \ref{fig:vis_sep_vggsound_2} present additional qualitative visualization of separating mixtures of two sound sources using V-SlowFast network from the MUSIC-21, AVE, and VGG-Sound datasets, respectively. Figure~\ref{fig:vis_sep_MUSIC21_3} and \ref{fig:vis_sep_MUSIC21_4} show results of separating mixtures of three and four sound sources from MUSIC-21.

\subsection{Sounding Source Localization}
\begin{figure*}[!thp]
    \centering
    \renewcommand\thefigure{G} 
    \includegraphics[width=1.0\linewidth]{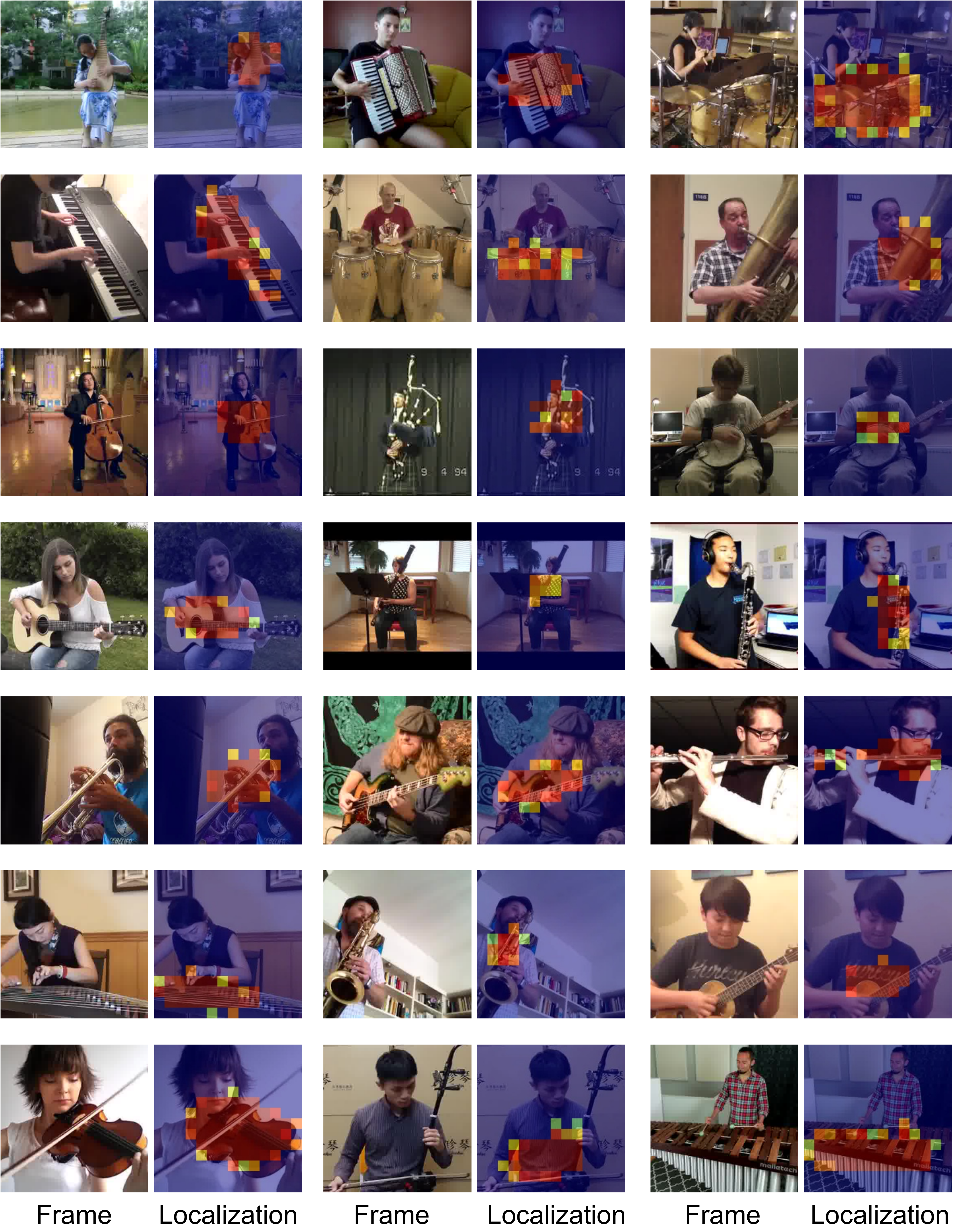}
   \caption{Visualization of the sound source localization using V-SlowFast network from MUSIC-21 dataset.}
\label{fig:vis_loc_MUSIC21}
\end{figure*}

\begin{figure*}[!thp]
    \centering
    \renewcommand\thefigure{H} 
    \includegraphics[width=1.0\linewidth]{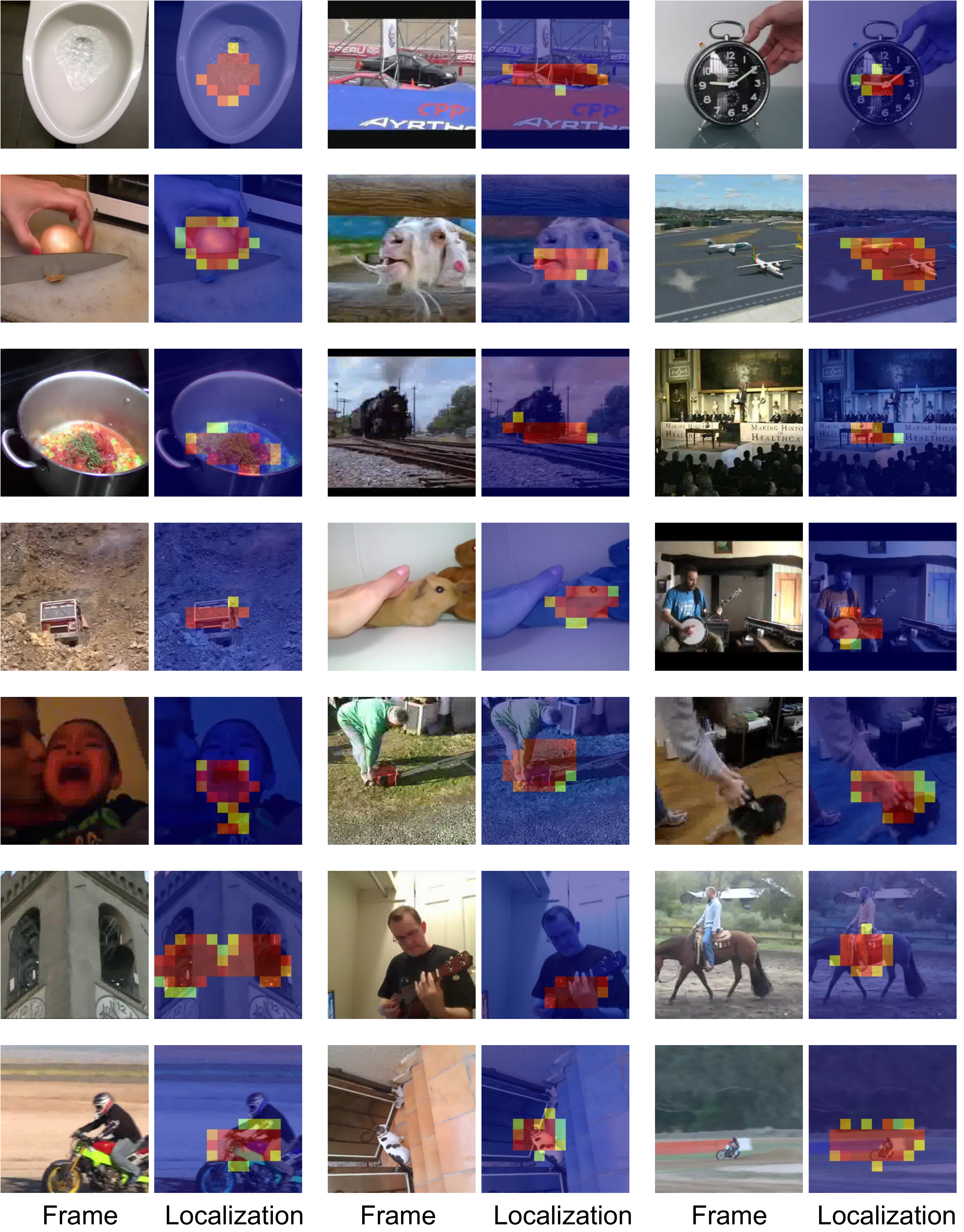}
   \caption{Visualization of the sound source localization using V-SlowFast network from AVE dataset.}
\label{fig:vis_loc_AVE}
\end{figure*}

\begin{figure*}[!thp]
    \centering
    \renewcommand\thefigure{I} 
    \includegraphics[width=1.0\linewidth]{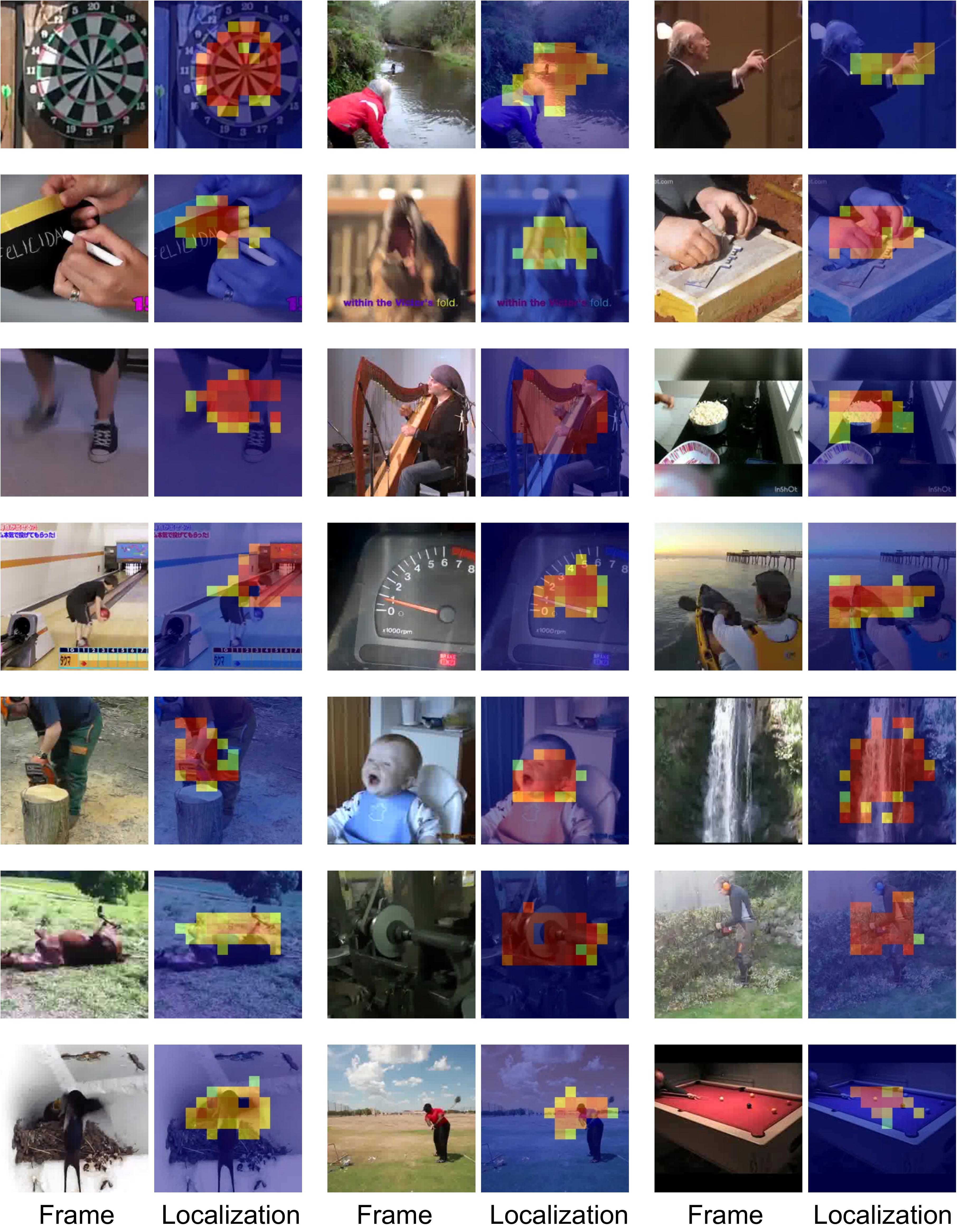}
   \caption{Visualization of the sound source localization using V-SlowFast network from VGG-Sound dataset.}
\label{fig:vis_loc_vggsound}
\end{figure*}

Figures~\ref{fig:vis_loc_MUSIC21}, \ref{fig:vis_loc_AVE} and \ref{fig:vis_loc_vggsound} provide additional qualitative visualization of the sound source localization with the proposed V-SlowFast framework using MUSIC-21, AVE, and VGG-Sound datasets, respectively.

\section{Implementation Details}
\label{sec:supp_implementation}

We extract video frames at 8 fps for all datasets and sub-sample audio signal at 11KHz, 22kHz, and 22KHz for MUSIC-21, AVE, and VGG-Sound datasets, respectively. We randomly crop 6-second audio clip and convert the input audio to F-T spectrogram using STFT with a hanning window of size 1022 (MUSIC-21, AVE) and 1498 (VGG-Sound), and a hop lengths of 256 (MUSIC-21), 184 (AVE) and 375 (VGG-Sound).  

A single frame ($224\times224$) is forwarded to the vision network. The vision network produces a compact representation $e_{v} \in \mathit{\mathbb{R}^{1 \times C_{V}}}$. $\textit{C}_{V}$ equals to $21$, $28$ and $310$ for MUSIC-21, AVE, and VGG-Sound datasets, respectively. The dimension of sound features $\textit{C}_{S}$ equals to $\textit{C}_{V}$, which represents the category numbers of dataset. 

$\alpha$ = 1 represents the full temporal resolution spectrogram. The slow spectrogram network and the fast spectrogram residual network take the low and high temporal resolution spectrograms as input, respectively. Thus, we consider $\alpha_{s} > 1$, and $\alpha_{f} < \alpha_{s}$. The $\phi^{inv}$ (inverse $\phi$) operation inverts the spectrogram into full temporal resolution spectrogram of $\alpha_{s}$ = 1 or $\alpha_{f}$ = 1. 

The proposed V-SlowFast model is implemented using Pytorch framework. We adopt stochastic gradient descent (SGD) with momentum 0.9, weight decay 1e-4, and batch size 10. The vision network, pre-trained on ImageNet, uses a learning rate of 1e-4, while all other of modules are trained from scratch using a learning rate of 1e-3.


\end{document}